\documentclass[11pt,oneside]{article}
\usepackage[letterpaper]{geometry}
\usepackage[dvipsnames]{xcolor}
\usepackage{graphicx}
\usepackage{amssymb,amsmath,amsthm}
\usepackage{float,placeins}
\usepackage{algorithm}
\usepackage{algpseudocode}
\usepackage{booktabs,mathtools}
\usepackage{siunitx}
\usepackage[numbers]{natbib}
\usepackage[hidelinks]{hyperref}

\newtheorem{lemma}{Lemma}[section]

\newtheorem{theorem}{Theorem}[section]

\theoremstyle{definition}
\newtheorem{definition}{Definition}[section]

\newtheorem{assumption}{Assumption}[section]
\theoremstyle{remark}
\newtheorem{remark}{Remark}

\newenvironment{keywords}{\par\smallskip\noindent\textbf{Keywords:}\enspace}{\par}
\newenvironment{MSCcodes}{\par\smallskip\noindent\textbf{MSC:}\enspace}{\par}
\newcommand{\email}[1]{\href{mailto:#1}{\texttt{#1}}}

\newcommand{\Shat}{\widehat{S}}
\newcommand{\shat}{\widehat{\sigma}}
\newcommand{\cB}{\mathcal{B}}
\newcommand{\cE}{\mathcal{E}}

\newcommand{\cK}{\mathcal{K}}
\newcommand{\cR}{\mathcal{R}}

\newcommand{\E}{\mathbb{E}}
\newcommand{\Prob}{\mathbb{P}}
\DeclareMathOperator*{\argmax}{argmax}

\title{Variance-Adaptive Upper Confidence Bounds for Multi-Armed Bandits
  with Kaplan--Meier Survival Estimation%
  \thanks{This work was partially supported by grants DMS-1907518, NSF/DMS 234961 and NSA H98230-24-1-0031.}}

\author{%
  Yangfan Cui\thanks{Department of Mathematics, North Carolina State University,
    Raleigh, NC 27695 (\email{ycui25@ncsu.edu}).}
  \and
  Andrew Papanicolaou\thanks{Department of Mathematics, North Carolina State University,
    Raleigh, NC 27695 (\email{apapani@ncsu.edu}).}
}

\begin{document}

\maketitle

%% ===========================================================================
%%  ABSTRACT
%% ===========================================================================
\begin{abstract}
  We study the stochastic multi-armed bandit problem in which the reward associated with each arm is the survival probability at a pre-specified time horizon, and the observed outcome is a right-censored time-to-event variable. Standard bandit algorithms face delayed and partial observations in this setting. Prior adaptations for the Kaplan--Meier (KM) estimator rely on Hoeffding-type bounds that ignore the inherent heteroscedasticity of continuous survival processes. To bridge this gap, we propose KM-UCB, a variance-adaptive policy that unites continuous-time survival analysis with discrete sequential decision-making. By leveraging martingale theory and continuous influence functions, we derive a fixed-cohort empirical Bernstein bound parameterized by Greenwood's variance formula. For realtime allocation, we retain the Greenwood index and use a survival bound from completed records to review the recommended arm. Under catch-up reporting, the policy has logarithmic expected regret with an arm-specific variance term in the exploration bound. We evaluate the original policy in a dual-layer delayed simulation and a retrospective replay of breast cancer trial data, and evaluate the correction in a separate paired simulation.
\end{abstract}



%% ===========================================================================
%%  1. INTRODUCTION
%% ===========================================================================
\section{Introduction}\label{sec:intro}

Multi-armed bandits (MABs) provide a framework for sequential decision-making under uncertainty \cite{auer2002finite,bubeck2012regret}. In clinical trials, MABs can guide adaptive treatment assignments so that more patients are allocated to superior therapies, while still gathering enough information to identify the best arm \cite{villar2015multi}. The ethical imperative in clinical settings is particularly acute: every suboptimal treatment assignment represents a real patient who may receive inferior care.

A distinguishing challenge in long-term trials is that outcomes are \emph{time-to-event} and often \emph{right-censored}: a patient's survival reward is delayed and may only be partially observed if the patient drops out or the trial ends before an event occurs. Standard bandit algorithms designed for immediate, uncensored rewards \cite{Chapelle2014} do not directly apply. This creates a fundamental mismatch between the classical MAB setting, where rewards are observed immediately after pulling an arm, and the survival analysis setting, where the reward may take months or years to materialize.

Consider, for example, a clinical trial comparing $K$ chemotherapy regimens for breast cancer, where the primary endpoint is 5-year disease-free survival. At any point during the trial, the algorithm must make arm selections despite having only partial and delayed feedback: due to staggered enrollment, some recently enrolled patients have not yet had enough follow-up time to reach the evaluation horizon (administrative censoring), while others may have dropped out early (lost to follow-up). Both mechanisms generate right-censored data, invalidating standard bandit bounds. An effective adaptive allocation algorithm must (i)~properly estimate each arm's survival probability from the available censored data, (ii)~quantify the uncertainty in these estimates, and (iii)~use this uncertainty to balance exploration and exploitation.

\subsection{Related work}\label{sec:related}

The multi-armed bandit framework has been extensively studied, with foundational work by Lai and Robbins~\cite{lai1985asymptotically} establishing asymptotic lower bounds for regret, and Auer, Cesa-Bianchi, and Fischer~\cite{auer2002finite} providing finite-time guarantees for UCB algorithms. Villar et al.~\cite{villar2015multi} provide a comprehensive overview of regret minimization in bandit algorithms applied to clinical trials, discussing UCB, Thompson Sampling, and Gittins index approaches.

In the context of clinical trials, most early bandit applications assumed rewards were observed after a fixed delay \cite{goel2010understanding}. However, these approaches do not fully address the complexities of survival analysis, where outcomes are time-to-event and often censored.

\paragraph{Censored bandits}
The integration of censored feedback into the multi-armed bandit framework was pioneered by \cite{NIPS2016_0bf727e9}, who paired Kaplan–Meier survival estimates with Hoeffding-type confidence bounds. Their algorithm operates in discrete epochs, recomputing KM estimates and eliminating arms whose upper bounds fall below the current best. While their framework is fundamentally sound and mathematically elegant, its Hoeffding-based confidence width scales purely with the sample size as $\mathcal{O}(\sqrt{\log t / n})$. In survival settings, however, estimation uncertainty is inherently heteroscedastic. Because the variance of the Kaplan--Meier estimator depends dynamically on the arm-specific censoring distribution (which depletes the at-risk population at different rates), replacing a generic sample-size bound with a censoring-aware variance proxy offers a natural pathway to more efficient exploration.

\paragraph{Bayesian approaches}
From a Bayesian perspective, Wu and Wager~\cite{wu2022partial} introduced Partial Likelihood Thompson Sampling (PL-TS), which adapts Thompson sampling to settings with delayed and censored feedback by updating posterior distributions using the Cox partial likelihood. Their framework enables efficient identification of superior treatments in vaccine trials with censored outcomes. However, PL-TS requires the proportional hazards assumption and provides average-case (Bayesian) guarantees rather than worst-case frequentist bounds, which are often preferred in safety-critical trial designs where regulatory agencies demand explicit control of Type I error rates.

\subsection{Contributions}\label{sec:contributions}

Departing from existing Bayesian strategies and conservative Hoeffding-type bounds, this paper introduces a frequentist approach that constructs data-adaptive confidence intervals via the Kaplan--Meier estimator and Greenwood's variance formula. The specific contributions of this paper are structured around both theoretical and empirical advances.

Our \textbf{KM-UCB} policy uses Kaplan--Meier survival estimates and Greenwood's variance formula. For fixed i.i.d.\ cohorts, Theorem~\ref{thm:km_concentration} establishes a continuous-time empirical Bernstein inequality using influence-function concentration and a Greenwood comparison.

For realtime allocation, we use an observable upper bound to review the arm recommended by the Greenwood index. Under the reporting model in Section~\ref{sec:setup}, the review policy has logarithmic expected regret with a variance-dependent exploration term (Theorem~\ref{thm:greenwood_regret}). Paired simulations assess the finite-horizon effect of the correction. The original dual-layer simulation and breast cancer analysis evaluate the original index.

\subsection{Organization}
Section~\ref{sec:km_ucb_concentration} establishes the core continuous-time concentration theory for the Kaplan--Meier estimator and Greenwood variance, including the influence-function decomposition and our main empirical Bernstein bound.
Section~\ref{sec:bandit_algorithm} introduces the survival bandit framework, the KM-UCB algorithm, and provides the instance-dependent regret analysis.
Section~\ref{sec:experiments} reports simulation and empirical results. Section~\ref{sec:conclusion} concludes.

%% ===========================================================================
%%  2. KM-UCB AND CONCENTRATION INEQUALITIES
%% ===========================================================================
\section{KM-UCB and concentration inequalities}
\label{sec:km_ucb_concentration}

\subsection{Survival analysis preliminaries}
\label{sec:background}

Let $\tau$ be a non-negative random variable representing the
time-to-event (e.g., time to death or relapse) with survival function
$S(t) = \Prob(\tau > t)$. Throughout this paper, we assume the distribution of $\tau$ is absolutely continuous with a probability density function $f(t)$. To formally describe the risk of event occurrence in a continuous-time framework, we define the \textit{instantaneous hazard rate} function $\lambda(t)$ as
\begin{equation}
  \lambda(t) = \lim_{\Delta t \to 0} \frac{\Prob(t \le \tau < t + \Delta t \mid \tau \ge t)}{\Delta t} = \frac{f(t)}{S(t)}.
\end{equation}
The \textit{cumulative hazard function} is naturally defined as the integral of the hazard rate, $\Lambda(t) = \int_0^t \lambda(u) du$, which allows the survival function to be expressed via the fundamental identity $S(t) = \exp(-\Lambda(t))$.
In practical clinical or bandit settings, the true survival time $\tau$ is often not fully observed due to right-censoring. Let $C$ denote the censoring time. We require the following standard assumption:

\begin{assumption}\textbf{(Independent censoring)}\label{ass:indep_cens}
  The survival time $\tau$ and censoring time $C$ are independent.
\end{assumption}

\subsubsection{The Kaplan--Meier estimator and Greenwood's formula}
\label{sec:km_greenwood}
To rigorously handle the time-to-event outcomes, we employ the standard continuous-time counting process framework \cite{FlemingHarrington1991}. For the $i$-th subject let $\tau_i$ denote the time-to-event and let $C_i$ denote the censoring time; the observed data for each subject is the pair $(X_i, \delta_i)$ with $X_i = \min(\tau_i, C_i)$ and $\delta_i = \mathbf{1}\{\tau_i \le C_i\}$, where $\delta_i = 1$ indicates an observed event and $\delta_i = 0$ indicates right-censoring. 
We define the individual counting process $N_i(t) = \mathbf{1}\{X_i \le t, \delta_i = 1\}$, which jumps by $1$ when the event is observed, and the at-risk process $Y_i(t) = \mathbf{1}\{X_i \ge t\}$, which indicates if the subject is under observation just prior to time $t$. Summing over the cohort yields the aggregated counting process $N(t) = \sum_{i=1}^n N_i(t)$ and the total at-risk process $Y(t) = \sum_{i=1}^n Y_i(t)$.
The empirical cumulative hazard function (Nelson-Aalen estimator) is defined via the stochastic integral
\begin{equation}
  \widehat{\Lambda}_n(t) = \int_0^t \frac{\mathbf{1}\{Y(u)>0\}}{Y(u)} dN(u).
\end{equation}
The Kaplan-Meier (KM) estimator \cite{Kaplan1958} is the nonparametric maximum likelihood estimator of $S(t)$, defined as the product-integral of the empirical hazard
\begin{equation}
  \widehat{S}_n(t) = \prod_{u \le t} \bigl(1 - d\widehat{\Lambda}(u)\bigr) = \prod_{u \le t} \left(1 - \frac{dN(u)}{Y(u)}\right),
\end{equation}
with $\widehat{S}_n(0) = 1$. Here, $dN(u)$ represents the number of events exactly at time $u$. The product-integral evaluates to the standard discrete product over the jump times of $N(u)$. The survival ratio $\widehat S_n(u)/S(u)$ is a martingale up to risk exhaustion. Appendix~\ref{app:km_properties} gives the precise statement.
The variance of the KM estimator is estimated using Greenwood's formula \cite{Greenwood1926}. While the theoretical variance integral for continuous distributions does not require a discrete correction, practical data often contain ties. To strictly account for empirical jumps, the Greenwood variance is defined as
\begin{definition}\textbf{(Greenwood's formula)}
The estimated variance of $\widehat{S}_n(t)$ is
\begin{equation}\label{eq:green_formula}
  \widehat{\sigma}_n^2(t) = \widehat{S}_n^2(t) \int_0^t \frac{d\widehat{\Lambda}(u)}{Y(u)\bigl(1 - \Delta\widehat{\Lambda}(u)\bigr)},
\end{equation}
where $\Delta\widehat{\Lambda}_n(u) = dN(u)/Y(u)$ is the jump size at time $u$.
\end{definition}
Note that in the completely uncensored setting, this formula gracefully reduces to the standard Bernoulli empirical variance $\Shat_n(t)(1-\Shat_n(t))/n$.

\subsubsection{Regularity assumptions}
The concentration results that follow require the following conditions on the hazard rates and the at-risk process, ensuring the survival process remains well-behaved over the horizon $[0, t]$.
\begin{assumption}\textbf{(Bounded hazard and at-risk probability)}
\label{assump:regularity}
There exist constants $p_{\min}\in(0,1]$ and $\Lambda_{\min}>0$
such that:
\begin{enumerate}
 \item For all $u\in[0,t]$, the population at-risk probability
 satisfies $y(u)=\mathbb P(X\ge u)\ge p_{\min}$.
 \item \textit{Non-degeneracy at the target horizon.} The cumulative
 hazard at the fixed target horizon satisfies $\Lambda(t)\ge\Lambda_{\min}$.
\end{enumerate}
\end{assumption}
\begin{remark}
The first condition is standard in survival studies with fixed follow-up, ensuring a sufficient proportion of the cohort remains under observation. Under absolute continuity, observed failures are almost surely untied, so the empirical hazard jump at a failure is $1/Y(u)$. Appendix~\ref{app:greenwood_compare} shows that $1-\Delta\widehat\Lambda_n(u)\ge1/2$ at every observed failure on the good event and for the stated sample size. This ensures that the denominators in Greenwood's formula are positive.
The non-degeneracy condition ensures that the asymptotic variance is bounded away from zero. Indeed, condition~1 gives $S(t)\ge y(t)\ge p_{\min}$. Since $y(u)\le1$, we obtain
\[
 V(t)=S^2(t)\int_0^t\frac{d\Lambda(u)}{y(u)}
 \ge S^2(t)\Lambda(t)\ge p_{\min}^2\Lambda_{\min}>0.
\]
\end{remark}

\subsection{Concentration inequalities for fixed sample size}
\label{sec:concentration_fixed}
\label{sec:concentration_known}

The KM estimator and Greenwood variance defined above are functions of a fixed cohort of $n$ i.i.d.\ censored observations from a single population.  In this section we establish their concentration properties in this fixed-sample setting, deferring the extension to random sample sizes (required by the bandit framework) to Section~\ref{sec:concentration_bandit}.\\
To establish finite-sample concentration inequalities for the Kaplan-Meier estimator in continuous time, our fundamental strategy is to decompose the estimation error into an empirical mean of independent influence functions and a negligible remainder.
\subsubsection{Influence-function decomposition}\label{sec:if_decomp}
\label{sec:concentration_greenwood}
Consider $n$ i.i.d.\ censored observations from a single population. Under the historical filtration $\mathcal{F}_t$, which encapsulates all event and censoring information up to time $t$, the compensated counting process $M_i(t) = N_i(t) - \int_0^t Y_i(u) d\Lambda(u)$ forms a zero-mean, square-integrable martingale \cite{FlemingHarrington1991}. This continuous-time martingale residual mathematically captures the random empirical fluctuation around the true hazard.

\begin{lemma}\textbf{(Influence-function decomposition of the KM estimator)}
\label{lem:IF_rep}
Under Assumptions~\ref{ass:indep_cens}--\ref{assump:regularity}, define the \emph{influence function} of $\Shat_n(t)$ \cite[Section~4.3]{Andersen1993} by
\begin{equation}\label{eq:IF_def}
  \widetilde{\psi}_{t}(Z_i)
  := -S(t)\int_0^{t} \frac{dM_i(u)}{y(u)},
\end{equation}
where $Z_i = (X_i, \delta_i)$.  Because $y(u)$ is a population-level (deterministic) quantity, the terms $\widetilde{\psi}_t(Z_1), \dots, \widetilde{\psi}_t(Z_n)$ are i.i.d.\ with $\E[\widetilde{\psi}] = 0$, and $\mathrm{Var}(\widetilde{\psi}) = V(t)$ the classical asymptotic variance of the KM estimator \cite{Gill1983,Andersen1993}.
Then the KM estimator based on $n$ i.i.d.\ samples satisfies 
\begin{equation}\label{eq:IF_decomp}
  \Shat_n(t) - S(t)
  = \underbrace{\frac{1}{n}\sum_{i=1}^n
    \widetilde{\psi}_{t}(Z_i)}_{L_n(t)}
  \;+\; \rho_n(t),
\end{equation}
where $|\rho_n(t)|\le C_r\log n/n$ almost surely for all $n\ge n_1$ along prefixes of a single i.i.d.\ sequence.
Here $C_r$ is deterministic and $n_1$ is an almost surely finite random starting index. Appendix~\ref{app:remainder_bound} gives an explicit finite-sample bound.

Moreover, the IF terms are bounded:
$|\widetilde{\psi}_{t}(Z_i)| \le b_\psi$ with $b_\psi = (1 + \Lambda(t))/p_{\min}$.
\end{lemma}

The influence-function representation and remainder bounds are proved in Appendices~\ref{app:if_proof} and~\ref{app:remainder_bound}.

\begin{theorem}\textbf{(Sub-Gaussian concentration of the KM estimator)}
\label{thm:KM-subGaussian}
Under Assumptions~\ref{ass:indep_cens}--\ref{assump:regularity}, there
is a constant $c_{\rm KM}>0$ depending only on $p_{\min}$ and
$\Lambda(t)$ such that for all $n\ge1$ and $\epsilon>0$,
\begin{equation}\label{eq:km_subgaussian}
 \Prob\bigl(|\Shat_n(t)-S(t)|>\epsilon\bigr)
 \le 2\exp(-c_{\rm KM}n\epsilon^2).
\end{equation}
\end{theorem}
\begin{proof}[Proof sketch]
The argument first controls the risk set at the fixed target. Because the risk set decreases with survival age, this also gives a lower bound at each earlier age. On this event the Nelson--Aalen error is a compensated counting-process integral. Its jumps and predictable variation give a two-sided tail bound. The Kaplan--Meier product differs from the exponential of the Nelson--Aalen estimate by an amount of order $1/n$ on the same event. This difference fits inside the deviation threshold when $n\epsilon^2$ is sufficiently large. For smaller $n\epsilon^2$, the probability bound is trivial. Appendix~\ref{app:fixed_concentration} gives the proof and the constant.
\end{proof}

In the sub-Gaussian bound of Theorem~\ref{thm:KM-subGaussian}, the tail decay is controlled by the cumulative hazard and the minimum at-risk probability. A radius built from these population bounds gives the same worst-case scale to samples whose actual variance can be quite different. In practice, the variance of the Kaplan--Meier estimator can be substantially smaller than this scale, particularly under mild or moderate censoring. Since the confidence radius governs exploration in the bandit, we now consider a variance-adaptive radius based on Greenwood's formula. Here the KM estimate and Greenwood variance are computed from $n$ i.i.d.\ observations.

Building upon the influence-function linearization in Lemma~\ref{lem:IF_rep}, the empirical Bernstein argument proceeds in two stages. First, Lemma~\ref{lem:gw_pop_equiv} compares the empirical Greenwood variance with the population variance $V(t)/n$ on a high-probability event. Second, Bernstein's inequality controls the average of the i.i.d.\ influence functions. The remainder bound in Section~\ref{app:remainder_bound} shows when the rest of the error fits inside the Greenwood radius. This yields the data-adaptive bound in Theorem~\ref{thm:km_concentration}.

\subsubsection{Greenwood variance and population variance}
\label{sec:gw_pop}

The asymptotic target of the Greenwood formula~\eqref{eq:green_formula} is $V(t)/n$. We compare the two quantities on a high-probability event.

\begin{definition}\textbf{(Good risk-set event)} \label{def:good_event}
We define the good event $\mathcal{R}_\delta$ as the intersection of the uniform concentration of the risk set and the cumulative hazard martingale, characterized by a tolerance parameter $\delta \in (0, p_{\min}/2)$:
\begin{equation} \label{eq:good_event}
  \mathcal{R}_\delta = \left\{ \sup_{u \le t} \left| \frac{Y(u)/n}{y(u)} - 1 \right| \le \delta \right\} \cap \left\{ \sup_{u \le t} |\widehat{\Lambda}_n(u) - \Lambda(u)| \le \delta \right\}.
\end{equation}
\end{definition}

\begin{lemma}\textbf{(Greenwood concentrates around population variance)}
\label{lem:gw_pop_equiv}
For $0<\delta<p_{\min}/2$, take $n_G(\delta)$, $C_G$ and $C_\Lambda$ as given in Appendix~\ref{app:greenwood_compare}.
For $n\ge n_G(\delta)$, on the event $\mathcal R_\delta$,
\[
 (1-C_G\delta)\frac{V(t)}{n}
 \le\widehat\sigma_n^2(t)
 \le(1+C_G\delta)\frac{V(t)}{n}.
\]
The Greenwood denominators are positive on this event. Moreover,
\begin{equation}\label{eq:good_event_prob}
 \mathbb P(\mathcal R_\delta^c)
 \le 2\exp(-2n\delta^2p_{\min}^2)
      +2\exp(-C_\Lambda n\delta^2).
\end{equation}
A strictly positive lower comparison requires $C_G\delta<1$.
\end{lemma}
Appendix~\ref{app:greenwood_compare} proves the comparison and the good-event probability bound.

\subsubsection{Main concentration theorem}\label{sec:main_conc}
\begin{theorem}\textbf{(Continuous-time empirical Bernstein bound for KM estimator)}
\label{thm:km_concentration}
Under Assumptions~\ref{ass:indep_cens}--\ref{assump:regularity}, fix $w\ge1$ and a tolerance $\delta$ satisfying the sample conditions in Appendix~\ref{app:fixed_concentration}. Then
\begin{align}
 &\Prob\!\left(|\Shat_n(t)-S(t)|>\sqrt{2w\shat_n^2(t)}\right)
 \notag\\
 &\qquad\le
 2e^{-c_\delta(n,w)w}+ \Prob(\mathcal R_\delta^c)+6e^{-w}.
 \label{eq:km_conc_gw}
\end{align}
The coefficient $c_\delta(n,w)>0$ is given by \eqref{eq:Rn_def}, and \eqref{eq:good_event_prob} bounds $\Prob(\mathcal R_\delta^c)$. Use Greenwood variance zero at an exhausted failure endpoint.
\end{theorem}
The fixed-cohort result applies to independent records. The decision-time analysis in Section~\ref{sec:bandit_algorithm} handles adaptive allocation and partial follow-up separately.

\section{The survival bandit model and KM-UCB algorithm}
\label{sec:bandit_algorithm}
\paragraph{Notation for the bandit setting}
We now introduce the multi-armed bandit framework. In Section 2, we utilized the subscript $n$ to denote the sample size in our empirical estimators (e.g., $\Shat_n(t)$). In the sequential setting, the sample size for arm $k$ up to trial round $T$ is a random variable. We denote this dynamic sample size (total pulls) by $N_k(T) = \sum_{j=1}^T \mathbf{1}\{a_j = k\}$. Furthermore, to track the right-censored feedback, we let $n_k(t,T)$ denote the number of patients in arm $k$ who remain at risk at survival time $t$ during trial round $T$.
In the sequential setting, the subscript tracks the number of assignments. Simultaneously, due to the delayed feedback, we do not suppress the dual-time arguments: we maintain both the continuous survival horizon $t$ and the discrete trial time $T$. Concretely, we define $\Shat_{N_k(T)}(t, T)$ and $\shat_{N_k(T)}^2(t, T)$.

These denote the Kaplan-Meier estimator and Greenwood variance evaluated at survival time $t$, computed using the reports available from arm $k$ at trial round $T$. 
Because our primary clinical endpoint is evaluated at a pre-specified target horizon $t^\dagger$, the bandit algorithm operates specifically on $\Shat_{N_k(T)}(t^\dagger, T)$ and $\shat_{N_k(T)}^2(t^\dagger, T)$. We denote the true survival probability and asymptotic variance at the target horizon for arm $k$ as $S_k(t^\dagger)$ and $V_k(t^\dagger)$, and we write $\Delta_k = S_{k^*}(t^\dagger) - S_k(t^\dagger)$ for the suboptimality gap.

\subsection{Survival bandit model}\label{sec:setup}\label{sec:setting}
Consider a stochastic MAB with $K$ arms indexed by $\mathcal K=\{1,\dots,K\}$ and a time horizon of $T$ rounds. At each trial round $j=1,\dots,T$, the learner selects an arm $a_j\in\mathcal K$ and a patient is enrolled.
For the patient enrolled at round $j$ and assigned to arm $a_j=k$, the true survival time $\tau_j^*$ and the independent censoring time $C_j$ are drawn from arm $k$'s underlying distributions. The latent censored pair is $(X_j,\delta_j)$, where $X_j=\min(\tau_j^*,C_j)$ and $\delta_j=\mathbf 1\{\tau_j^*\le C_j\}$. Each arm has a sequence of i.i.d.\ potential records in assignment order, independent across arms and of policy randomization. Survival times are continuous. For the regret analysis, assume a common lower bound $p>0$ on the probability of $X_j\ge t^\dagger$ under each arm's record law.

\paragraph{Reporting and partial follow-up.}
Patient $i$ enters the observation pool at round $i+D_i$. The delays $D_i$ are independent geometric random variables with parameter $r\in(0,1)$ on $\{1,2,\ldots\}$, independent of the potential records and policy randomization. Upon entry, the learner receives all follow-up accrued since enrollment and thereafter receives follow-up as it accrues. For an entered patient, the record visible at round $T$ is
\[
 X_i(T)=\min\{X_i,T-i\},\qquad
 \delta_i(T)=\delta_i\mathbf 1\{X_i\le T-i\}.
\]
A target-horizon record is complete once a terminal event or censoring by $t^\dagger$ is reported, or follow-up through $t^\dagger$ is available. Reports distinguish actual censoring from temporary end of follow-up. Completed records are retained. The decision at round $j$ uses only the assignments and reports through $j-1$ and policy randomization.

Because patient intake is an ongoing process, the KM estimation at trial time $T$ incorporates all available follow-up on each arm. Let $\mathcal I_k(T)=\{j\le T:a_j=k\}$ denote the index set of patients assigned to arm $k$ up to round $T$. The total pull count is exactly the cardinality of this set, $N_k(T)=|\mathcal I_k(T)|$. To maintain a strictly continuous-time formulation, we avoid discrete-time grid approximations of survival age. The at-risk process for arm $k$ during trial round $T$ is
\begin{equation}\label{eq:realtime_risk}
 n_k(u,T)=
 \sum_{\substack{i\in\mathcal I_k(T)\\i+D_i\le T}}
     \mathbf 1\{X_i(T)\ge u\}, \qquad u\ge0.
\end{equation}
A failure at age $u$ contributes to the event count only if its visible record has $X_i(T)=u$ and $\delta_i(T)=1$. The KM estimate and Greenwood variance use these counts in the formulas of Section~\ref{sec:km_greenwood}, evaluated at $t^\dagger$.

The reward of arm $k$ is its survival probability at the pre-specified target horizon $t^\dagger$: $\mu_k=S_k(t^\dagger)$. Let $k^*$ denote an optimal arm and $\Delta_k=S_{k^*}(t^\dagger)-S_k(t^\dagger)$ the suboptimality gap.

\begin{definition}\textbf{(Cumulative regret)}\label{def:regret}
The cumulative (pseudo-)regret after $T$ rounds is
\begin{equation}\label{eq:regret_def}
  \cR(T)=\sum_{j=1}^T
       \bigl(S_{k^*}(t^\dagger)-S_{a_j}(t^\dagger)\bigr)
       =\sum_{k\ne k^*}\Delta_k N_k(T).
\end{equation}
In expectation, $\E[\cR(T)]=\sum_{k\ne k^*}\Delta_k\,\E[N_k(T)]$. This performance measure remains a purely theoretical construct based on true survival probabilities, whereas the actual policy uses the observed reports and the estimates computed from them.
\end{definition}

\subsection{The KM-UCB algorithm}\label{sec:algorithm}
We first present the algorithm under the idealized assumption that a variance proxy $V_k(t^\dagger)$ is known for each arm.  This simplified setting clarifies the structure of the exploration bonus before we introduce the Greenwood plug-in.

\begin{definition}\textbf{(KM-UCB with known variance)}\label{def:KMUCB_known}
After a warm-start phase that pulls each arm $n_{\mathrm{init}}$ times, the arm choice following trial time $T$ is
\begin{equation}\label{eq:ucb_known}
  a_{T+1} = \argmax_{k \in \cK} \left[
    \Shat_{N_k(T)}(t^\dagger,T) +
    \sqrt{\frac{2\alpha\,V_k(t^\dagger)\,\ln T}{N_k(T)}}
  \right],
\end{equation}
where $\alpha > 1$ is the exploration parameter. Here $V_k(t^\dagger)$ is a deterministic variance proxy. The decision-time conditions for this benchmark are stated in Section~\ref{sec:regret_known}.
\end{definition}

In practice, a suitable variance proxy is unknown and must be estimated from the heavily censored data.
The Greenwood plug-in gives the following data-driven, variance-adaptive index.

\begin{definition}\textbf{(KM-UCB with Greenwood variance)}\label{def:KMUCB}
At trial round $j$, the original index is
\begin{equation}\label{eq:ucb_gw}
 \mathrm{UCB}_k(j)=
 \widehat S_{N_k(j-1)}(t^\dagger,j-1)
 +\sqrt{2\alpha\,\widehat\sigma_{N_k(j-1)}^2(t^\dagger,j-1)
             \ln(j-1)}.
\end{equation}
Here $\widehat\sigma_{N_k(j-1)}^2(t^\dagger,j-1)$ is the Greenwood variance from~\eqref{eq:green_formula} and $\alpha>1$. Both quantities use all reports through round $j-1$. Use index zero if no record is available. At an exhausted failure endpoint, take the Greenwood variance to be zero. In the appendix, $U_k^0(t)=\mathrm{UCB}_k(t+1)$ is shorthand for this index.
\end{definition}

\subsubsection{A review step using completed records}
Let $C_k^{\mathrm p}(t)$ be the upper confidence bound computed from the longest completed prefix of patients assigned to arm $k$, in assignment order. A record is complete if a terminal outcome is reported or if follow-up through $t^\dagger$ has been observed. Section~\ref{app:exponential_certificate} gives the calculation.

Let $b_j$ be the index maximizer and $k_j$ the maximizer of $C_k^{\mathrm p}(j-1)$. Let $\ell_k(j-1)$ be the latest round through $j-1$ in which arm $k$ was pulled. The policy selects $b_j$ unless the following two conditions hold, in which case it selects $k_j$:
\begin{equation}\label{eq:review_rule}
 C_{k_j}^{\mathrm p}(j-1)>
 \mathrm{UCB}_{b_j}(j)+c\,\frac{\ln(j-1)}{N_{b_j}(j-1)},
 \qquad
 j-1-\ell_{k_j}(j-1)\ge L .
\end{equation}
The review parameters $c\ge0$ and $L$ are fixed. The first condition asks whether the candidate's survival bound warrants reconsidering the original recommendation. The second gives the candidate a waiting period between pulls.

Algorithm~\ref{alg:KMUCB} gives the complete procedure.

\begin{algorithm}[t]
\caption{KM-UCB with Greenwood variance and a review step}
\label{alg:KMUCB}\label{alg:certificate}
\begin{algorithmic}[1]
  \Require Arms $\mathcal K$, horizon $T$, target $t^\dagger$, $\alpha>1$,
  $n_{\mathrm{init}}\ge1$, calibration $\gamma\ge2$, review parameters
  $c\ge0$ and integer $L\ge0$.
  \State \textbf{Initialization:} Pull each arm $n_{\mathrm{init}}$ times
  in the first $J_0=K n_{\mathrm{init}}$ rounds, cycling through the arms.
  \State Record the latest pull $\ell_k(J_0)$ of each arm during initialization.
  \For{trial round $j=J_0+1,\ldots,T$}
    \State Observe the newly available outcomes through round $j-1$,
    including accrued records of newly entered patients and updates to
    previously reported follow-up.
    \For{each arm $k\in\mathcal K$}
      \State Update the dynamic risk set $n_k(u,j-1)$ and event
      count through age $t^\dagger$.
      \State Compute the KM estimate and Greenwood variance from all
      available records, then form $\mathrm{UCB}_k(j)$ by \eqref{eq:ucb_gw}.
      \State Extend the completed prefix and compute $C_k^{\mathrm p}(j-1)$.
    \EndFor
    \State Find $b_j$ and $k_j$, breaking ties by arm order.
    \State Set $a_j=b_j$. If both conditions in \eqref{eq:review_rule}
    hold, set $a_j=k_j$.
    \State Assign the $j$-th patient to arm $a_j$ and set its latest pull to $j$.
  \EndFor
  \State Process the reports through round $T$.
  \State \Return the final KM estimate at $t^\dagger$ for each arm with reports.
\end{algorithmic}
\end{algorithm}

\begin{remark}\textbf{(Comparison with Censored-UCB \cite{NIPS2016_0bf727e9})}
A Hoeffding-type confidence width uses the pull count. The Greenwood bonus uses the risk sets at the observed failure times, so its value can differ across arms with the same number of pulls. The review step uses the complete-prefix bound for its optimism check.
\end{remark}

\subsection{Concentration with random sample sizes}
\label{sec:concentration_bandit}
The fixed-sample results of Section~\ref{sec:concentration_fixed} assume a deterministic sample size $n$. In the bandit setting $N_k(T)$ is random, since it depends on the learner's past decisions. Some of these assignments have only partial reports by round $T$. To pass from fixed samples to the bandit estimator, we work first with the completed prefix of each possible size, then compare it with the estimator computed from all available reports.

\begin{theorem}\textbf{(Concentration for KM-bandit estimator)}
\label{thm:bandit_concentration}
Under the survival bandit model in Section~\ref{sec:setup}, fix $T\ge2$ and $\alpha>1$. Let $V_{\mathrm{sub},k}=1/c_{\mathrm{KM},k}$, with the sub-Gaussian constant from Theorem~\ref{thm:KM-subGaussian} evaluated for arm $k$. Write $n=n_k^{\mathrm{eff}}(T)$ for the length of its completed prefix. For $n\ge8x_T/p$, define
\[
 \cB_{k,n}(T)=
 \sqrt{\frac{V_{\mathrm{sub},k}x_T}{n}}
 +\frac{2h_T^{(\alpha)}}{np},
\]
where $x_T=(\alpha+1)\ln T+\ln(3K+1)$ and the reporting allowance $h_T^{(\alpha)}$ is given in Section~\ref{app:bandit_concentration_proof}. With probability at least $1-T^{-\alpha}$,
\begin{equation}\label{eq:bandit_conc}
 \left|\Shat_{N_k(T)}(t^\dagger,T)-S_k(t^\dagger)\right|
 \le \cB_{k,n}(T)
\end{equation}
for every arm satisfying $n=n_k^{\mathrm{eff}}(T)\ge8x_T/p$. On the same event $n_k^{\mathrm{eff}}(T)\ge N_k(T)-h_T^{(\alpha)}$ for all arms.
\end{theorem}

\begin{proof}[Proof sketch]
Apply the fixed-sample sub-Gaussian result to each deterministic prefix, take a union bound, and then use the actual completed prefix. This gives the square-root term. Adding the remaining reports can change the KM product through its risk sets and its additional failure times. The comparison with the prefix gives the second term. Section~\ref{app:bandit_concentration_proof} gives the probability calculation and the reporting allowance.
\end{proof}

\subsubsection{Greenwood-based UCB and the review bound}
At each round, the Greenwood index first recommends an arm using all follow-up reported so far. To decide whether to reconsider that recommendation, we compute a one-sided survival bound for each arm from its completed prefix. A high bound can make an arm a candidate for review, subject to the waiting condition in \eqref{eq:review_rule}. In particular, the best arm's bound can keep it eligible for review even when its original Greenwood index is low. The following theorem controls the probability that a true survival probability lies above its review bound.

\begin{theorem}\textbf{(Complete-prefix upper confidence bound)}
\label{thm:certificate}
For every $t\ge2$, simultaneously over arms, the actual adaptively chosen complete prefixes satisfy
\[
 \mathbb P\bigl(\exists k:C_k^{\mathrm{p}}(t)<\mu_k\bigr)\le t^{1-\gamma}.
\]
\end{theorem}
Appendix~\ref{app:exponential_certificate} proves the result by covering every deterministic prefix before using the observed prefix.

\subsection{Regret analysis}\label{sec:regret}
Write $V_k(t^\dagger)$ for the asymptotic Kaplan--Meier variance of arm $k$ at the target age.

\subsubsection{Regret with known variance proxy}\label{sec:regret_known}
We first state the regret bound for the idealized known-variance setting. 
\begin{theorem}\textbf{(Regret bound with known variance proxy)}
\label{thm:sublinear_regret}
Assume $\Delta_k>0$ for all $k\ne k^*$ and consider KM-UCB with the known variance proxy in Definition~\ref{def:KMUCB_known}, with $\alpha>1$. At decision $j=t+1$, write $\beta_k(t)=\sqrt{2\alpha V_k(t^\dagger)\ln t/N_k(t)}$. Suppose the probability of either one sided error
\[
 \Shat_{N_{k^*}(t)}(t^\dagger,t)+\beta_{k^*}(t)<S_{k^*}(t^\dagger)
 \quad\text{or}\quad
 \Shat_{N_k(t)}(t^\dagger,t)>S_k(t^\dagger)+\beta_k(t)
\]
is at most $A_\alpha t^{-\alpha}$ for each $k\ne k^*$ and $t\ge J_0$, where $A_\alpha$ is independent of $t$ and $T$. Then the expected cumulative regret satisfies
\begin{equation}\label{eq:regret_known}
 \E[\cR(T)]\le\sum_{k\ne k^*}\Delta_k\left(
 \frac{8\alpha\,V_k(t^\dagger)\ln T}{\Delta_k^2}
 +n_{\mathrm{init}}+C_\alpha'\right),
\end{equation}
where $C_\alpha'=1+A_\alpha/(\alpha-1)$ is independent of $T$. In particular, $\E[\cR(T)]=O(\ln T)$.
\end{theorem}
\begin{proof}[Proof sketch]
At a decision where neither estimation error occurs, selecting arm $k$ implies $\Delta_k\le2\beta_k(t)$. Squaring gives the exploration threshold. At decisions selecting $k$, each pull count before the decision occurs at most once. Add the initial pulls and sum the error probabilities by decision round. Multiplying by the gaps and summing proves \eqref{eq:regret_known}.
\end{proof}
The error assumption is for the specific index in Definition~\ref{def:KMUCB_known}. The realtime radius in
Theorem~\ref{thm:bandit_concentration} includes a reporting term and does not by itself establish this assumption. The Greenwood policy below obtains decision-time control through the review rule.

\subsubsection{Regret with Greenwood variance}\label{sec:regret_gw}

We now derive the variance-adaptive regret bound for the Greenwood plug-in version. The analysis follows the standard UCB decomposition \cite{bubeck2012regret} and accounts for the review step. Let $k^*$ be the optimal arm. At a decision that selects a suboptimal arm, we consider three events
\begin{align*}
 \cE_1(j)&:\ \text{the complete-prefix bound for the optimal arm is too low},\\
 \cE_2(j)&:\ \text{an index or certificate exceeds its proof bound},\\
 \cE_3(j)&:\ \text{an exploration width is at least half the selected arm's gap}.
\end{align*}
If a review is blocked by the waiting condition, the width in $\cE_3(j)$ can belong to the review candidate instead of the selected arm. Section~\ref{app:three_events} gives the precise events.

\begin{lemma}\textbf{(Decomposition of expected pulls)}\label{lem:decomp}
If $a_j=k$ for a suboptimal arm $k$ at a decision $j>J_0$, then at least one of $\cE_1(j)$, $\cE_2(j)$, or $\cE_3(j)$ holds. Consequently,
\begin{align*}
 \E[N_k(T)]\le n_{\mathrm{init}}
 &+\sum_{j=J_0+1}^{T}
 \Prob\bigl((\cE_1(j)\cup\cE_2(j))\cap\{a_j=k\}\bigr)\\
 &+\sum_{j=J_0+1}^{T}\Prob\bigl(\cE_3(j)\cap\{a_j=k\}\bigr).
\end{align*}
\end{lemma}

\begin{proof}[Proof sketch]
When the first two events fail, the optimal arm's certificate covers its survival probability. If the review is executed or the original recommendation passes the comparison, selection of a suboptimal arm requires a sufficiently large width. When the review is blocked, the selected arm's index is at least the candidate's original index. Comparing the two indices and the candidate's certificate gives the third event. Section~\ref{app:three_events} proves the three comparisons and bounds their contributions.
\end{proof}

\begin{lemma}\textbf{(Bound on estimation failures)}\label{lem:est_fail}
The gap-weighted contribution of $\cE_1$ and $\cE_2$ satisfies
\[
 \sum_{j=J_0+1}^T
 \E\!\left[\Delta_{a_j}\mathbf1_{\cE_1(j)\cup\cE_2(j)}\right]
 \le 3\ln T .
\]
\end{lemma}
\begin{proof}[Proof sketch]
The bounds for completed prefixes, reporting and index widths hold simultaneously at a given decision except on an event of probability at most $3(j-1)^{1-\gamma}$. Summing over decision rounds counts every affected decision, including repeated low estimates based on the same optimal-arm data. Details are in Section~\ref{app:three_events}.
\end{proof}

\begin{lemma}\textbf{(Bound on exploration steps)}\label{lem:explore}
Outside $\cE_1$ and $\cE_2$, the gap-weighted contribution of $\cE_3$ is at most $B_{\mathrm{sel}}(T)+B_{\mathrm{wait}}(T)$, given in \eqref{eq:selected_review_cost} and \eqref{eq:blocked_review_cost}. Both terms are $O(\ln T)$ for fixed review parameters and a fixed bandit instance.
\end{lemma}
\begin{proof}[Proof sketch]
After selection of a suboptimal arm, its pull count increases. This bounds the number of decisions that can be attributed to its large width. When a blocked review is attributed to the candidate, its most recent pull can account for only $L$ decisions. Section~\ref{app:three_events} gives the two counts and their gap-weighted bounds.
\end{proof}

Combining the lemmas yields our main regret theorem.

\begin{theorem}\textbf{(Regret bound for KM-UCB with Greenwood variance)}
\label{thm:greenwood_regret}
Under the survival bandit model of Section~\ref{sec:setup} with $K\ge2$, assume there is a unique optimal arm $k^*$. For Algorithm~\ref{alg:KMUCB} with fixed review parameters $c$ and $L$, there are constants $C$ and $C_{\mathrm{rem}}$, independent of $T$, such that for $T\ge J_0$,
\begin{equation}\label{eq:sharpened_regret}
 \mathbb E[\mathcal R(T)] \le
 C\sum_{k\ne k^*}\frac{V_k(t^\dagger)}{\Delta_k}\ln T
 + C_{\mathrm{rem}}\ln T .
\end{equation}
Section~\ref{app:regret_review} specifies the constants and the contribution of the review waiting period.
\end{theorem}
The term $V_k(t^\dagger)/\Delta_k$ gives the explicit variance dependence of the exploration contribution. The remainder includes reporting delays and blocked reviews and can also depend on the gaps.
\begin{proof}[Proof sketch]
Cover the optimal arm by its complete-prefix bound and count confidence failures by decision round. On the remaining rounds, a suboptimal selection requires an index width comparable to its gap. When a review is blocked, the difference between the candidate's upper bound and original index may account for part of that gap. Charge these rounds to the candidate's latest pull, which can account for at most $L$ decisions. All other exploration is counted by the number of pulls of the selected arm. The width bounds and the full counting argument are in Section~\ref{app:regret_review}.
\end{proof}

\section{Numerical experiments}\label{sec:experiments}
\subsection{Simulation setup}\label{sec:sim_setup}
Each arm $k \in \{0,1,2\}$ generates survival times from an exponential distribution with hazard rate $\lambda_k$. For the \emph{mid-gap} configuration, we assign $(\lambda_0, \lambda_1, \lambda_2) = (0.12, 0.15, 0.162)$. At the horizon $t^\dagger = 5$, these parameters imply baseline survival probabilities of roughly $(0.55, 0.47, 0.45)$. The gaps are $\Delta_0 = 0$ (optimal), $\Delta_1 \approx 0.08$, $\Delta_2 \approx 0.10$. These gaps are nontrivial but meaningful, making the learning problem neither trivial nor degenerate.

\paragraph{Choice of survival threshold}
We fix $t^\dagger = 5$. If $t^\dagger$ is too small, all arms have survival probabilities near one and the differences vanish. If $t^\dagger$ is too large, all probabilities approach zero. At $t^\dagger = 5$, the survival probabilities are sufficiently separated to provide meaningful learning signals, while remaining clinically interpretable (e.g., a ``5-year survival rate'').

\paragraph{Censoring mechanism}
Censoring times follow $\mathrm{Exp}(\nu)$ with $\nu = 0.18$, plus an administrative cutoff at $c_{\max} = 10$.

\paragraph{Dual-layer delay mechanism}
A distinctive feature of our simulation is a dual-layer delay mechanism that accurately reflects the practical realities of clinical trials by integrating both intrinsic and extrinsic delays. The first layer consists of an intrinsic, endogenous delay: the survival indicator $\mathbf{1}\{\tau_k^* > t^\dagger\}$ cannot be observed immediately, as the system must inherently wait until either the event occurs or the evaluation threshold $t^\dagger$ is reached. During this observational window, patients with partial follow-up contribute to the Kaplan--Meier estimate up to their observed follow-up age. 

Beyond this natural biological waiting time, we introduce a second layer of extrinsic, exogenous delay to account for administrative friction, such as data entry lags and centralized reporting bottlenecks. We model this administrative delay using a geometric random variable $D \sim \mathrm{Geom}(p)$ with $p = 0.6$. Consequently, a patient enrolled at trial round $j$ does not immediately update the algorithm's information state; rather, their continuous-time trajectory only enters the available observation pool at round $j + D$.

\paragraph{Evaluation metrics} We evaluate performance using four key metrics: the cumulative regret $\mathcal{R}(T) = \sum_{j=1}^T (S_{k^*}(t^\dagger) - S_{a_j}(t^\dagger))$, the Greenwood variance $\widehat{\sigma}_{N_k(T)}^2(t^\dagger, T)$ for each arm to quantify estimation stability, the arm pull counts $N_k(T)$ to illustrate sampling allocation dynamics over time, and the best-arm selection ratio $\rho(T) = N_{k^*}(T)/\sum_{k \in \mathcal{K}} N_k(T)$.

\subsection{Baselines}\label{sec:baselines}
We benchmark KM-UCB against four algorithmic baselines.

\paragraph{Censored-UCB (2016) \cite{NIPS2016_0bf727e9}}
Uses the KM estimator with a Hoeffding-style confidence width $\sqrt{2\alpha\ln j / \max\{1, N_k(j)\}}$. We adapt the original epoch-based algorithm to a fully sequential version for fair comparison.

\paragraph{Naive-UCB}
Treats survival as a standard Bernoulli bandit by discarding all censored observations. It isolates the subset of patients with \emph{determinate} outcomes by round $j$ (i.e., those who either reached $t^\dagger$ or experienced an event before $t^\dagger$), and computes the simple empirical success rate $\widehat{S}_k^{\mathrm{Bern}}$. The confidence bound is $\mathrm{UCB}_k(j) = \widehat{S}_k^{\mathrm{Bern}} + \sqrt{\alpha\ln j / \tilde{N}_k(j)}$, where $\tilde{N}_k(j)$ is the dynamically available number of determinate samples.

\paragraph{$\varepsilon$-Greedy}
Explores uniformly with probability $\varepsilon_j = \varepsilon_0/ \sqrt{j}$; otherwise exploits the arm with the highest Bernoulli-proxy survival estimate.

\paragraph{Uniform}
Selects each arm with equal probability at every round.

\subsection{Simulation results}\label{sec:sim_results}

The original Greenwood index in \eqref{eq:ucb_gw} is used in the baseline comparisons below.
These comparisons use $T = 1000$ rounds, $n_{\mathrm{init}} = 6$ warm-start samples per arm, $\alpha = 1.2$, and are averaged over 50 random seeds.

\paragraph{Arm allocation}
Table~\ref{tab:sim_summary} summarizes the allocation statistics. KM-UCB allocates an average of 641.9 patients to the optimal arm (Arm~0), achieving a best-arm pull ratio of 0.642. $\varepsilon$-Greedy and Censored-UCB rank second and third at 0.530 and 0.520, while Naive-UCB and Uniform achieve only 0.443 and 0.331, respectively.

\begin{table}[t]
\centering
\caption{Simulation results: mean arm pulls and best-arm ratio ($T=1000$, 50 seeds).}\label{tab:sim_summary}
\sisetup{table-format=3.1}
\begin{tabular}{l SSS S[table-format=1.3]}
\toprule
\textbf{Policy} & {Arm 0 (Opt)} & {Arm 1} & {Arm 2} & {Best-arm ratio} \\
\midrule
KM-UCB (Greenwood)     & 641.9 & 201.3 & 156.8 & 0.642 \\
$\varepsilon$-Greedy   & 530.5 & 229.8 & 239.7 & 0.530 \\
Censored-UCB (2016)    & 519.7 & 247.2 & 233.1 & 0.520 \\
Naive-UCB              & 442.7 & 290.6 & 266.7 & 0.443 \\
Uniform                & 331.2 & 332.9 & 336.0 & 0.331 \\
\bottomrule
\end{tabular}
\end{table}
A noteworthy observation is that KM-UCB exhibits slightly larger standard errors across random seeds, particularly for the optimal arm. This reflects greater \emph{adaptivity} to stochastic censoring patterns rather than instability in the estimator. Specifically, small differences in early censoring realizations lead to distinct exploration trajectories.

\begin{figure}[t]
  \centering
  \includegraphics[width=\linewidth]{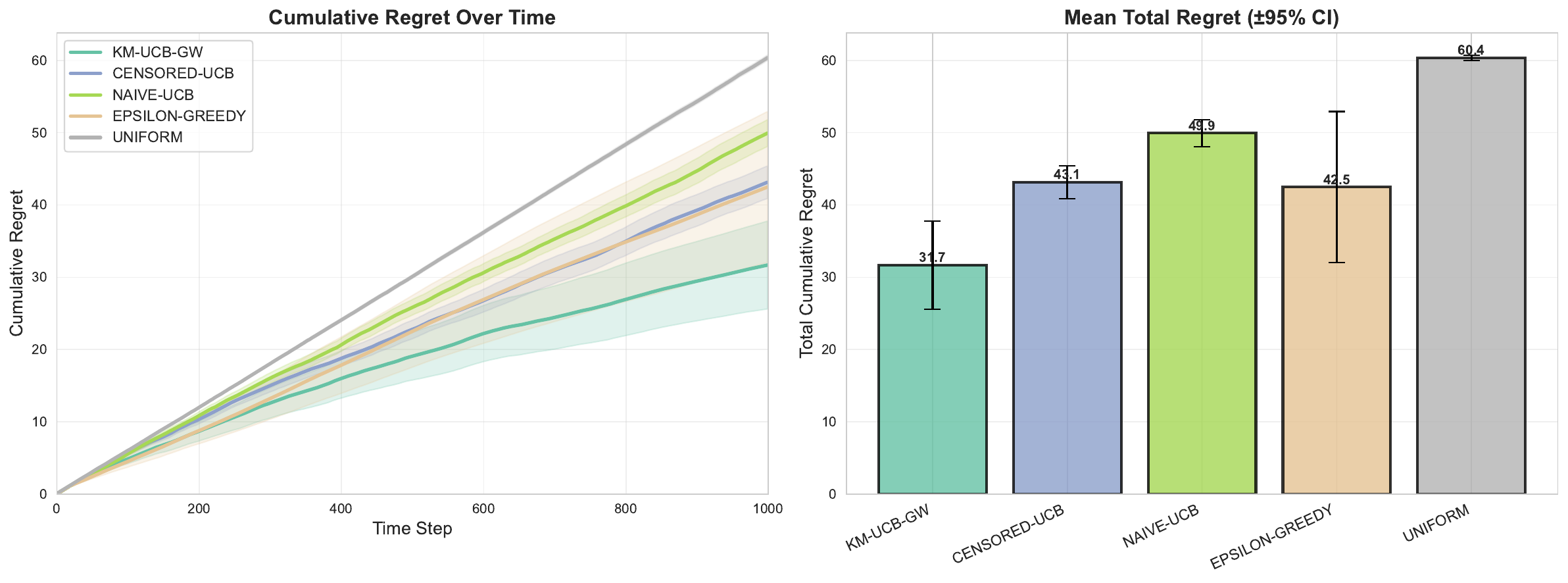}
  \caption{Cumulative regret of all algorithms over $T=1000$ rounds ($K=3$ arms, mid-gap configuration, exponential censoring with $\nu=0.18$, geometric feedback delay with $p=0.6$). Shaded regions indicate mean $\pm 1.96$ SE across 50 seeds. KM-UCB (Greenwood) achieves the lowest mean regret at the end of the simulation.}
  \label{fig:regret}
\end{figure}

\paragraph{Cumulative regret}
Figure~\ref{fig:regret} displays the cumulative regret curves. At $T=1000$, KM-UCB has the lowest mean cumulative regret (31.7), followed by $\varepsilon$-Greedy (42.5), Censored-UCB (43.1), Naive-UCB (49.9), and Uniform (60.4).

\begin{figure}[t]
  \centering
  \includegraphics[width=\linewidth]{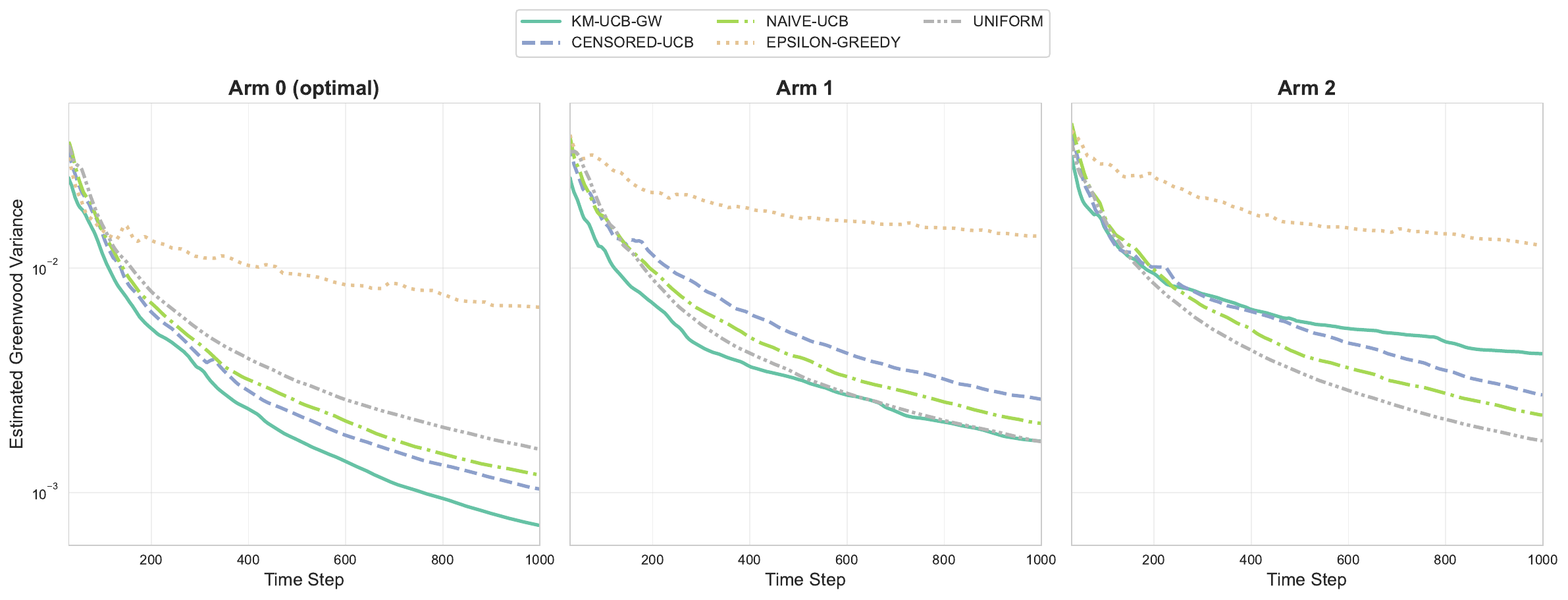}
  \caption{Estimated Greenwood variance $\widehat{\sigma}_{N_k(T)}^2(t^\dagger, T)$ for each arm under different algorithms, averaged over 50 seeds. Lower variance indicates more stable KM estimation.}
  \label{fig:gw_var}
\end{figure}

\paragraph{Greenwood variance trajectories}
Figure~\ref{fig:gw_var} shows the estimated Greenwood variance over time. For the optimal arm (Arm~0), KM-UCB achieves the lowest variance, reflecting its heavier allocation to this arm. The ordering differs across the suboptimal arms. By the end of the simulation, the KM-UCB estimate is among the lowest for Arm~1 and is higher than all but $\varepsilon$-Greedy for Arm~2.

\begin{figure}[t]
  \centering
  \includegraphics[width=\linewidth]{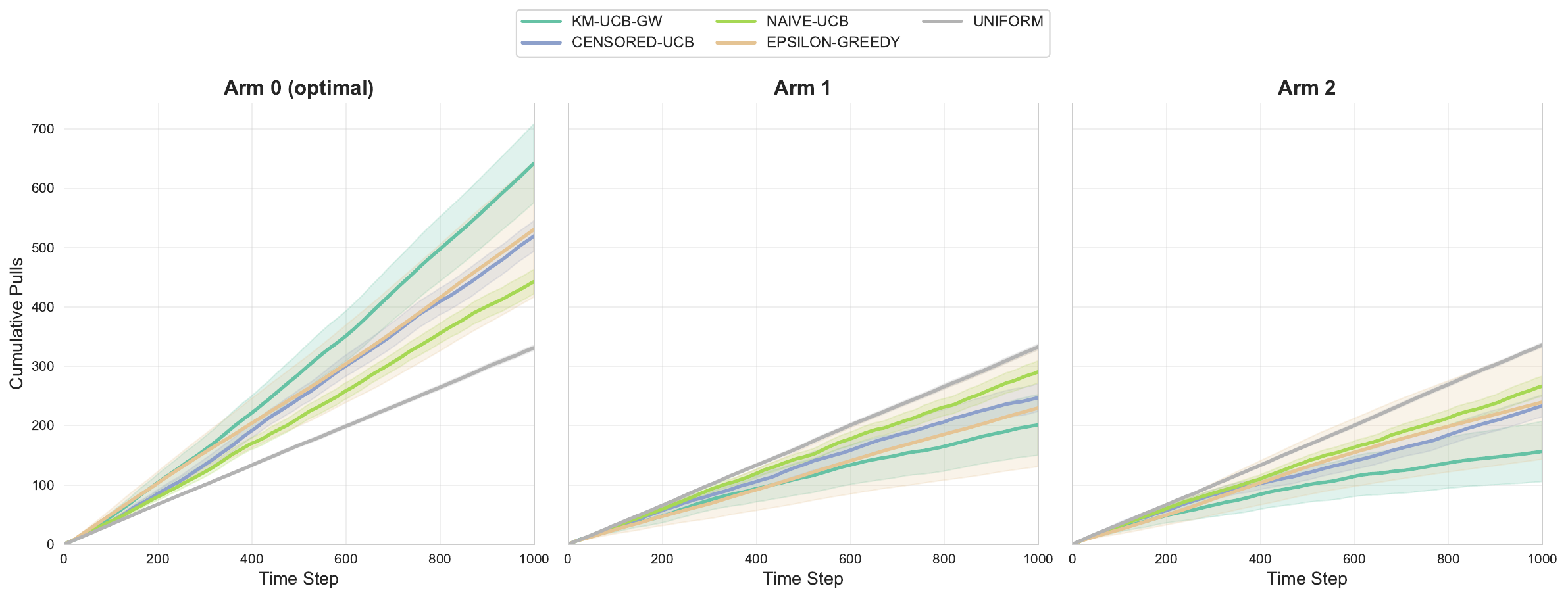}
  \caption{Arm pull counts $N_k(T)$ across algorithms. Each panel shows one arm; shaded regions indicate mean $\pm 1.96$ SE across 50 seeds.}
  \label{fig:arm_pulls}
\end{figure}

\paragraph{Arm pull dynamics}
Figure~\ref{fig:arm_pulls} compares the cumulative arm pulls over time. KM-UCB quickly concentrates on the optimal arm (Arm~0), while the number of pulls on suboptimal arms stabilizes at much lower levels. Naive-UCB and $\varepsilon$-Greedy distribute pulls more evenly, with $\varepsilon$-Greedy exhibiting high variance due to persistent exploration episodes.

\begin{figure}[t]
  \centering
  \includegraphics[width=0.5\linewidth]{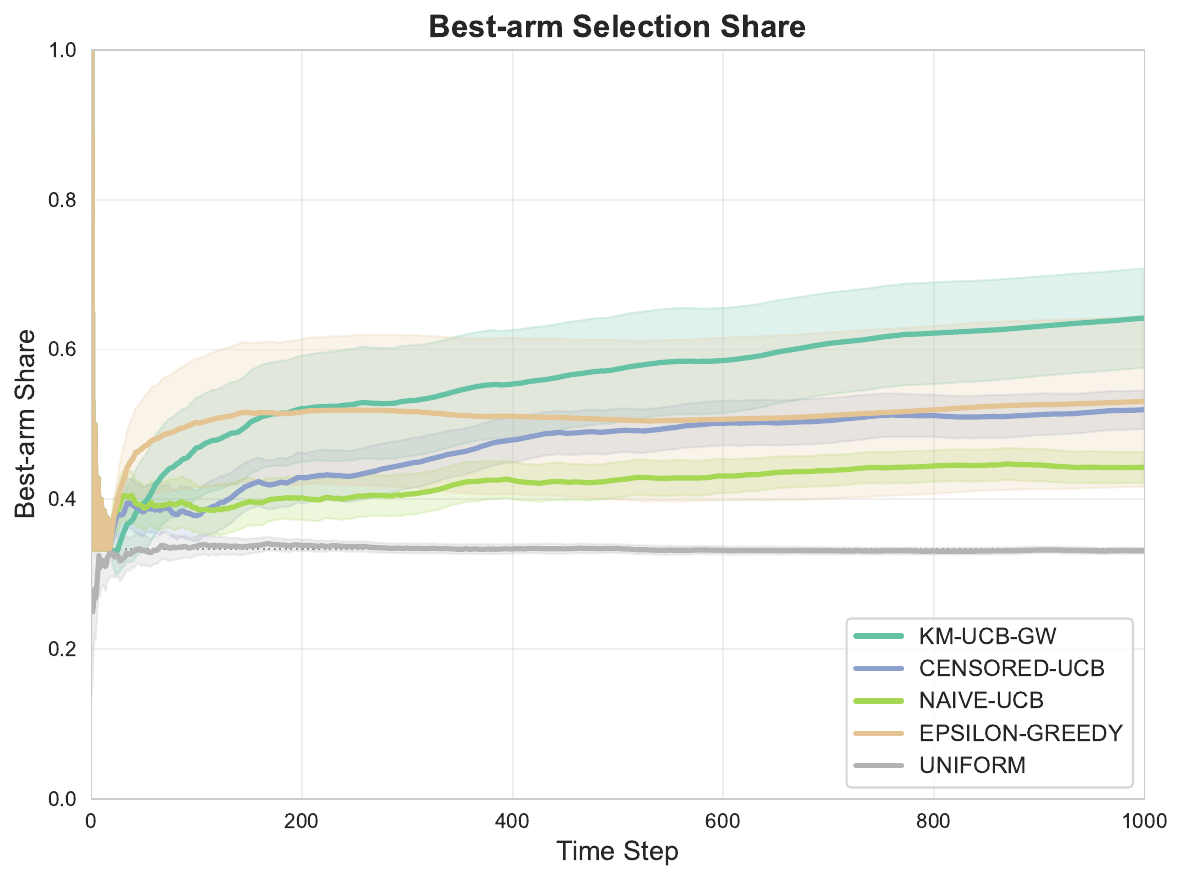}
  \caption{Selection share of the best arm $N_{k^*}(T)/T$ under five policies. Solid lines: mean across 50 seeds; shaded bands: mean $\pm 1.96$ SE.}
  \label{fig:best_share}
\end{figure}

\paragraph{Best-arm selection share}
Figure~\ref{fig:best_share} tracks the proportion of patients allocated to the best arm over time. KM-UCB converges to the highest selection share, reaching approximately 0.64 by $T = 1000$.

\FloatBarrier
\paragraph{Paired comparison of the review rule.}
\label{sec:new_validation}
Using the event and censoring distributions of Section~\ref{sec:sim_setup}, we compare the original Greenwood index with the review rule under the catch-up reporting model in Section~\ref{sec:setup}. We set $T=5000$, use six initial pulls per arm, and set $\alpha=1.2$, $\gamma=2$ and $c=64$. The policies share latent event and censoring streams, calendar delays and randomized arm labels on seeds 100--199. The original policy uses \eqref{eq:ucb_gw} alone. The review policies use the same complete-prefix bound and differ in their waiting condition.

\begin{table}[htbp]
\centering
\caption{Paired review experiments under catch-up reporting. Standard errors of mean regret are in parentheses. A run is stalled when the original policy never revisits the best arm after initialization. The last column counts reviews that overrode an original recommendation of the true best arm, summed over all runs.}
\label{tab:catchup_validation}
\begin{tabular}{@{}lccc@{}}
\toprule
Policy & Mean regret &
 \shortstack{Stalled runs with at least\\1,000 later best-arm pulls} &
 \shortstack{Overrides of\\the best arm}\\
\midrule
Original KM-UCB & $105.50$ ($11.08$) & $0/9$ & $0$\\
Review, $L=0$ & $92.14$ ($2.97$) & $9/9$ & $5{,}907$\\
Review, $L=100$ & $91.39$ ($3.76$) & $9/9$ & $1{,}081$\\
Review, $L=250$ & $94.26$ ($5.04$) & $8/9$ & $477$\\
Review, $L=500$ & $96.07$ ($6.16$) & $8/9$ & $268$\\
\bottomrule
\end{tabular}
\end{table}

All four review policies revisited the best arm in each of the nine stalled runs. Compared with review without a waiting period, $L=100$ reduced overrides of the best arm by $81.7\%$ and retained all nine recoveries according to the criterion in Table~\ref{tab:catchup_validation}. The paired regret difference was $-0.76$ (SE $1.36$) relative to review with $L=0$ and $-14.12$ (SE $8.55$) relative to original KM-UCB. On seed 115, the best arm was pulled 2,311, 193 and 8 times after initialization for $L=100$, 250 and 500, respectively. The waiting periods are compared on these same seed pairs.

\FloatBarrier
\subsection{Empirical evaluation: breast cancer trial}\label{sec:breast}
This evaluation uses the original Greenwood index in \eqref{eq:ucb_gw}.

\subsubsection{Dataset description}
We evaluate KM-UCB on a real-world breast cancer clinical trial dataset.\footnote{This clinical trial data was obtained from the National Cancer Institute via Project Data Sphere in June 2024. Project Data Sphere and the National Cancer Institute have not contributed to or approved of the work presented in this paper.} The trial studied two chemotherapy backbones (CA and T) and two treatment durations (4 vs.\ 6 cycles), yielding four arms: $\mathcal{A} = \{\text{CA-4}, \text{CA-6}, \text{T-4}, \text{T-6}\}$. The primary endpoint is disease-free survival (DFS), with time measured in months and event indicator from censoring status ($1 = $ event, $0 = $ censored). Overall survival is used only for sensitivity checks.

\subsubsection{Retrospective replay protocol}
We perform $N = 500$ random permutations of the patient order under a fixed survival threshold $t^\dagger \in \{36, 60, 96\}$ months. At each trial round $j$, a policy assigns an arm $a_j \in \mathcal{A}$ and updates its statistics upon observing $(X_j, \delta_j)$. An oracle arm $a^*(t^\dagger)$ is defined using arm-wise KM on the full cohort: $a^*(t^\dagger) = \argmax_a \widehat{S}_a(t^\dagger)$, with ties broken by the restricted mean survival time $\mathrm{RMST}_a(t^\dagger) = \int_0^{t^\dagger} \widehat{S}_a(u) du$. Right-censored observations contribute to KM through the risk set up to their censoring time. In Bernoulli-proxy baselines (Naive-UCB, $\varepsilon$-Greedy, Uniform), only determinate outcomes are used: $X_j \ge t^\dagger$ counts as success, $\delta_j = 1$ with $X_j < t^\dagger$ counts as failure, and observations censored before $t^\dagger$ are excluded.

\subsubsection{Results}
\begin{figure}[t]
  \centering
  \includegraphics[width=\linewidth]{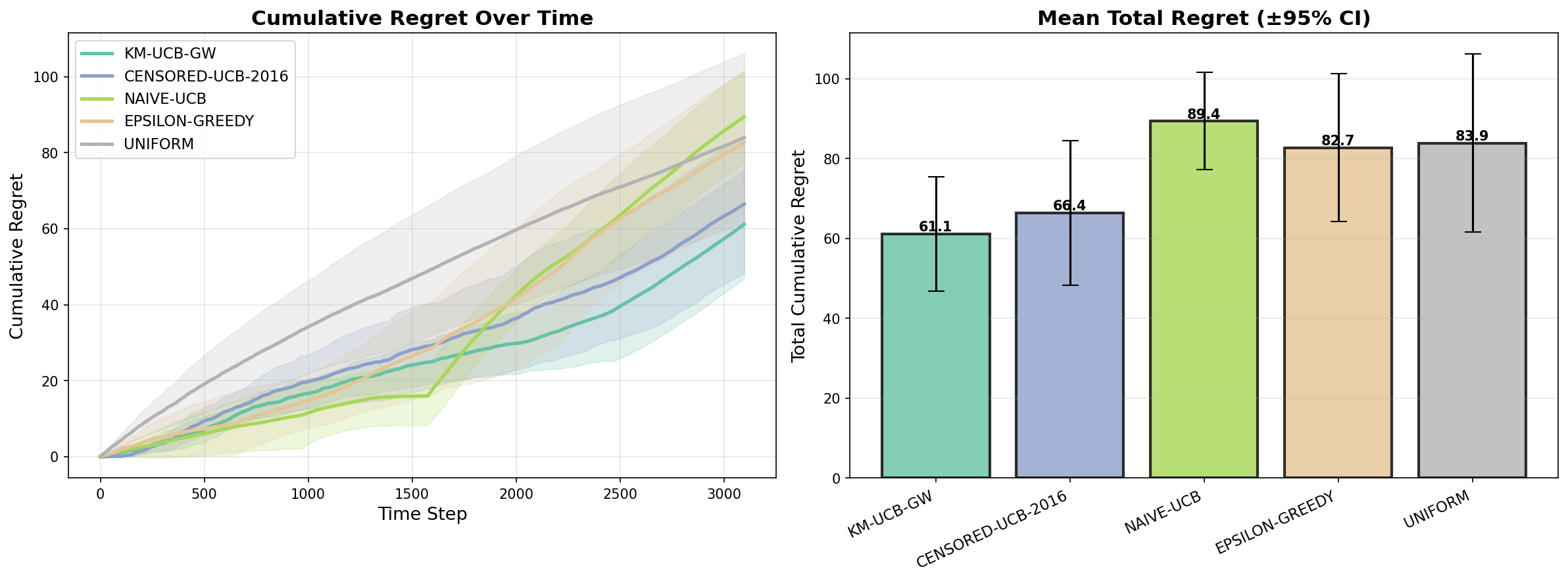}
  \caption{Comparison of adaptive allocation policies on the breast cancer dataset. Left: cumulative regret over patient index (500 permutations, $\pm 95\%$ CI). Right: total regret summary.}
  \label{fig:breast_regret}
\end{figure}

At the end of the replay, KM-UCB has the lowest mean cumulative regret (61.1), followed by Censored-UCB (66.4), $\varepsilon$-Greedy (82.7), Uniform (83.9), and Naive-UCB (89.4), as shown in Figure~\ref{fig:breast_regret}. The mean for KM-UCB is approximately 8\% lower than that of Censored-UCB.

\paragraph{Why KM-based methods dominate}
The experimental results reveal several critical insights regarding algorithmic performance under severe censoring. Fundamentally, KM-based methods significantly outperform proxy-based approaches like Naive-UCB because the Kaplan--Meier estimator leverages the complete survival trajectory, actively incorporating all available samples regardless of their censoring status. In contrast, Naive-UCB relies on a Bernoulli proxy that simply discards censored observations. This structural weakness is severely amplified under high censoring regimes, where the proxy method's effective sample size is drastically reduced, leading to high estimation variance and an irreversible sample depletion that prevents the algorithm from correcting early mistakes.

Furthermore, within the class of KM-based methods, incorporating the Greenwood variance yields substantial additional gains. KM-UCB has a lower mean final regret than Censored-UCB in this replay. Its Greenwood-derived confidence width adapts to the specific censoring pattern of each arm. By contrast, the Hoeffding-based width of Censored-UCB treats the uncertainty of all arms identically, resulting in an overly conservative exploration radius that fails to capitalize on the heteroscedastic nature of the survival data.

%% ===========================================================================
%%  5. CONCLUSION
%% ===========================================================================

\section{Conclusion}\label{sec:conclusion}
We have proposed KM-UCB, a variance-adaptive bandit algorithm for sequential treatment allocation with right-censored survival outcomes. Greenwood's estimate sets the original exploration bonus. Reviewing the recommended arm gives logarithmic expected regret under the specified reporting model. In the paired simulation, the waiting condition reduced overrides of the best arm while retaining recovery in the runs where the original policy stalled.

Several promising directions naturally emerge from our framework. Theoretically, developing a minimax lower bound that explicitly accounts for the underlying censoring structure is a critical next step to formally establish the optimality of our derived regret rate. Practically, extending KM-UCB to incorporate patient covariates via semiparametric frameworks, such as Cox proportional hazards models or a specialized Cox-UCB algorithm, would enable personalized treatment allocation. Finally, adapting this variance-aware approach to handle non-proportional hazards or integrating non-stationary censoring dynamics remains an important and challenging frontier for real-world clinical trials.

%% ===========================================================================
%%  ACKNOWLEDGMENTS
%% ===========================================================================
\section*{Acknowledgments}
The breast cancer clinical trial data was obtained from the National Cancer Institute via Project Data Sphere.  Project Data Sphere and the National Cancer Institute have not contributed to or approved of the work presented in this paper.

%% ===========================================================================
%%  REFERENCES
%% ===========================================================================

\clearpage
\setcounter{section}{0}
\renewcommand{\thesection}{S.\arabic{section}}
\begin{center}
{\Large Supplementary Materials}\\[6pt]
Variance-Adaptive Upper Confidence Bounds for Multi-Armed Bandits\\
with Kaplan--Meier Survival Estimation\\[6pt]
Yangfan Cui and Andrew Papanicolaou
\end{center}
\section{Counting Process Formulation}\label{app:counting_process}
To rigorously analyze estimators with time-varying risk sets, we record each subject through the observed pair $(X_i,\delta_i)$, where $X_i=\min(\tau_i,C_i)$ and $\delta_i=\mathbf 1\{\tau_i\le C_i\}$. The associated counting-process objects are defined directly from this observed data \cite{Andersen1993,FlemingHarrington1991}.

\begin{definition}\textbf{(Counting process)}
The counting process 
\[
  N_i(t)=\mathbf 1\{X_i\le t,\ \delta_i=1\}
\]
records whether the event for subject $i$ has been observed by time $t$. It is a right-continuous step function that jumps from $0$ to $1$ exactly at an uncensored event time.
\end{definition}

\begin{definition}\textbf{(At-risk process)}
The at-risk process 
\[
  Y_i(t)=\mathbf 1\{X_i\ge t\}
\]
indicates whether subject $i$ is still under observation just prior to time $t$.
\end{definition}

A central result is the Doob--Meyer decomposition: assuming the hazard rate $\lambda(t)$ exists, the counting process decomposes as
\begin{equation}\label{eq:doob_meyer}
  N_i(t) = \underbrace{\int_0^t Y_i(u)\lambda(u)\,du}_{\text{compensator } \Lambda_i(t)} + \underbrace{M_i(t)}_{\text{martingale}},
\end{equation}
where $M_i(t)$ is a zero-mean local square-integrable martingale with respect to the natural filtration. This martingale property is the engine that drives our concentration proofs since it allows us to apply Freedman's inequality and related tools to the censored setting.

\section{Properties of the KM Estimator and Greenwood Variance}\label{app:km_properties}

\begin{lemma}\textbf{(Survival ratio before risk exhaustion)}
\label{lem:KM_unbiased}
Under independent censoring the survival ratio $\widehat S_n(u)/S(u)$
stopped at risk exhaustion is a martingale on any fixed interval
where $S$ is bounded away from zero.
\end{lemma}
\begin{proof}
Up to exhaustion, the product formula gives
$d(\widehat S_n/S)=-(\widehat S_n/S)_{-}\,d\bar M/Y$, where $(\cdot)_{-}$ represents the left limit.
The stopped ratio is bounded on the stated interval and hence is a
martingale. After exhaustion the estimator stays constant, while
$S(u)$ can continue decreasing. For example, with one subject
$\E\widehat S_1(t)=S(t)+\Prob(C<\tau\le t)$.
\end{proof}

\begin{lemma}\textbf{(Greenwood variance upper bound)}\label{lem:gw_upper}
Provided that $Y(t) > 0$, the empirical Greenwood variance $\widehat{\sigma}_n^2(t)$ is upper-bounded by:
\begin{equation}\label{eq:gw_upper_fixed}
  \widehat{\sigma}_n^2(t) \le \frac{\widehat S_n^2(t)}{n} \left(\frac{n}{Y(t)} - 1\right).
\end{equation}
\end{lemma}
\begin{proof}
The distinct pre-event risk counts are integers greater than $Y(t)$,
almost surely, since there is no failure at the fixed target time.
Hence the variance sum is bounded by
$\sum_{y=Y(t)+1}^n [y(y-1)]^{-1}=1/Y(t)-1/n$.
Multiplication by $\widehat S_n^2(t)$ proves the inequality.
\end{proof}

\begin{lemma}\textbf{(Greenwood variance lower bound)}\label{lem:gw_lower}
At a fixed time $t$ where the Greenwood denominators are positive,
\begin{equation}\label{eq:gw_lower_fixed}
  \widehat{\sigma}_n^2(t) \ge \frac{\widehat S_n^2(t)}{n^2}\, \mathbf{1}\{N(t) \ge 1\}.
\end{equation}
\end{lemma}
\begin{proof}
The Greenwood variance is given by the integral $$\widehat{\sigma}_n^2(t) = \widehat S_n^2(t) \int_0^t \frac{dN(u)}{Y(u)(Y(u)-dN(u))}$$. Because the total initial risk set is $n$, the at-risk process satisfies $Y(u) \le n$ for all $u \ge 0$. Consequently, each denominator in the integral is strictly bounded above by $n^2$. We can then lower-bound the integral by factoring out the maximum denominator
\[
  \int_0^t \frac{dN(u)}{Y(u)(Y(u)-dN(u))} \ge \int_0^t \frac{dN(u)}{n^2} = \frac{N(t)}{n^2}.
\]
Since the event count $N(t)$ is a non-negative integer, we naturally have $N(t) \ge \mathbf{1}\{N(t) \ge 1\}$. Multiplying both sides by $\widehat S_n^2(t)$ yields the desired lower bound.
\end{proof}

\section{Proof of the Influence-Function Decomposition}
\label{app:if_proof}
Write $p=p_{\min}$, $h=\Lambda(t)$ and $\bar M=\sum_i M_i$.
By the martingale isometry and independence across subjects,
$L_n=-S(t)\int_0^t(ny(u))^{-1}d\bar M(u)$ has centered i.i.d.\ summands
with variance $S(t)^2\int_0^t y(u)^{-1}d\Lambda(u)=V(t)$.
Each individual martingale has at most one observed event, and its
compensator through $t$ is at most $h$. Thus
$|\widetilde\psi_t|\le(1+h)/p$.

On $\inf_{u\le t}Y(u)\ge np/2$ with $np/2\ge2$, the Nelson--Aalen
error is $\widehat\Lambda_n(t)-h=\int_0^t Y^{-1}d\bar M$. Define
$Q_n=-\log\widehat S_n(t)-\widehat\Lambda_n(t)$. At the distinct
event times, the integer risks exceed $s=Y(t)$ almost surely.
By $-\log(1-z)-z\le z^2/(1-z)$ and telescoping,
$0\le Q_n\le\sum_{y=s+1}^n1/[y(y-1)]\le2/(np)$.
For $d=\log\widehat S_n(t)-\log S(t)$, Taylor's theorem for the
exponential between the two nonpositive log-survival values gives
$|\widehat S_n-S-Sd|\le d^2/2$. Subtracting $L_n$ leaves a remainder
of magnitude at most
\[
 \left|\int_0^t\left(Y(u)^{-1}-(ny(u))^{-1}\right)d\bar M(u)\right|
 +\frac{2}{np}+\frac12\left(
 |\widehat\Lambda_n(t)-h|+\frac2{np}\right)^2 .
\]
S.4 bounds the two stochastic integrals and supplies the deterministic
radius needed in the finite-sample results.

\section{Proof of the Remainder Bound}\label{app:remainder_bound}
For $0<x\le np^2/2$ let $a=\sqrt{x/(2n)}$. DKW gives
$\sup_{u\le t}|Y(u)/n-y(u)|\le a$ outside probability $2e^{-x}$.
On this event $Y(u)\ge np/2$ and
$|Y(u)^{-1}-(ny(u))^{-1}|\le b=2a/(np^2)$.
Clip this predictable difference to $[-b,b]$, taking $1/Y$ as zero
at zero risk. The clipped integral has jumps bounded by $b$ and
bracket bounded by $b^2nh$. For the hazard error use the predictable
integrand $\mathbf 1\{Y\ge np/2\}/Y$, with jump bound $2/(np)$
and bracket bound $2h/(np)$. Both integrals agree with their
unclipped counterparts on the DKW event. The two-sided
Bernstein--Freedman inequality \cite{Freedman1975} includes the
linear jump terms. Outside an additional probability $4e^{-x}$,
the first integral has magnitude at most
$b\sqrt{2nhx}+2bx/3$ and the second at most
$2\sqrt{hx/(np)}+4x/(3np)$. For $n\ge4/p$, combining these two bounds
with the Taylor inequality in Appendix~\ref{app:if_proof} gives
\begin{align}
 \eta_n(x)
 &=b\sqrt{2nhx}+\frac{2bx}{3}+\frac2{np}\notag\\
 &\quad+\frac12\left(2\sqrt{\frac{hx}{np}}+
                         \frac{4x}{3np}+\frac2{np}\right)^2,\notag\\
 \Prob(|\rho_n(t)|>\eta_n(x))&\le6e^{-x}.
 \label{eq:full_remainder_tail}
\end{align}
The remainder is defined everywhere by $\widehat S_n-S-L_n$, including
when the risk set is exhausted. For $x=2\log n$ the displayed radius is
$O((\log n)/n)$ and the failure probabilities are summable. Along
prefixes of an infinite i.i.d.\ sequence Borel--Cantelli gives the
almost sure rate with an almost surely finite starting index.

\section{Proof of Lemma~\ref{lem:gw_pop_equiv} (Greenwood concentration)}\label{app:greenwood_compare}
\begin{proof}
We consider a fixed cohort of $n$ i.i.d.\ observations. Write
$p=p_{\min}$, $\ell=\Lambda_{\min}$, and $h=\Lambda(t)$.
Assume $0<\delta<p/2$ and
\begin{equation}\label{eq:n_G_threshold}
 n\ge n_G(\delta):=
 \left\lceil\frac{4(h+1)}{p\delta}\right\rceil.
\end{equation}
As in the main text, let
\[
 \mathcal A_\delta=
 \left\{\sup_{u\le t}
 \left|\frac{Y(u)/n}{y(u)}-1\right|\le\delta\right\},\qquad
 \mathcal R_\delta=
 \mathcal A_\delta\cap
 \left\{\sup_{u\le t}
 |\widehat\Lambda_n(u)-\Lambda(u)|\le\delta\right\}.
\]

\paragraph{Survival and hazard on the good event.}
Absolute continuity of the event distribution implies that each
observed event is a single jump, almost surely. On $\mathcal R_\delta$,
\[
 Y(u)\ge np(1-\delta)\ge np/2,\qquad
 \Delta\widehat\Lambda_n(u)\le\frac{2}{np}\le\delta.
\]
Thus Greenwood's denominators are positive on this event.
For $0\le z<1$, $0\le-\log(1-z)-z\le z^2/(1-z)$.
Consequently,
\begin{equation}\label{eq:log_S_bound}
 0\le-\log\widehat S_n(t)-\widehat\Lambda_n(t)
 \le
 \frac{\max_{u\le t}\Delta\widehat\Lambda_n(u)}
      {1-\max_{u\le t}\Delta\widehat\Lambda_n(u)}
 \widehat\Lambda_n(t)
 \le \frac{4(h+1)}{np}\le\delta.
\end{equation}
Here $\widehat\Lambda_n(t)\le h+\delta\le h+1$ was used.
It follows that
$|\log\widehat S_n(t)-\log S(t)|\le2\delta$ and hence, with
$q_n=\widehat S_n(t)^2/S(t)^2$,
\[
 |q_n-1|\le4e^2\delta,\qquad q_n\le e^2.
\]

\paragraph{Comparison of the variance integrals.}
Define
\[
 G_n=\int_0^t
  \frac{d\widehat\Lambda_n(u)}
       {(Y(u)/n)(1-\Delta\widehat\Lambda_n(u))},
 \quad
 J_n=\int_0^t\frac{d\widehat\Lambda_n(u)}{y(u)},
 \quad
 P=\int_0^t\frac{d\Lambda(u)}{y(u)}.
\]
On $\mathcal R_\delta$,
$r(u)=(Y(u)/n)/y(u)\in[1-\delta,1+\delta]$ and
$z(u)=\Delta\widehat\Lambda_n(u)\in[0,\delta]$.
For $0<\delta\le1/2$ these bounds imply
\[
 \left|\frac{1}{r(u)(1-z(u))}-1\right|\le6\delta.
\]
Therefore
\[
 |G_n-J_n|\le6\delta J_n
 \le\frac{6(h+1)}p\,\delta.
\]
For the remaining term, put
$F_n(u)=\widehat\Lambda_n(u)-\Lambda(u)$ and $f(u)=1/y(u)$.
The function $f$ is nonnegative, nondecreasing and bounded by $1/p$.
Integration by parts for functions of bounded variation gives
\[
 |J_n-P|
 =\left|\int_0^t f(u)\,dF_n(u)\right|
 \le\bigl(f(t)+\operatorname{TV}_{[0,t]}(f)\bigr)
       \sup_{u\le t}|F_n(u)|
 \le\frac{2\delta}{p}.
\]
The bound holds with the left-limit convention for $f$ even if the
censoring distribution has atoms. Equivalently, one can represent
$f$ as a mixture of terminal-interval indicators; every such interval
has $dF_n$-mass bounded in absolute value by $2\delta$.
Since $P\ge\Lambda(t)\ge\ell>0$, we obtain
\[
 \left|\frac{G_n}{P}-1\right|
 \le D_G\delta,\qquad
 D_G=\frac{6h+8}{p\ell}.
\]
Using $n\widehat\sigma_n^2(t)/V(t)=q_nG_n/P$ now yields
\begin{equation}\label{eq:C_G_explicit}
 \left|\frac{n\widehat\sigma_n^2(t)}{V(t)}-1\right|
 \le C_G\delta,\qquad
 C_G=e^2\left(4+\frac{6h+8}{p\ell}\right).
\end{equation}
This proves the two-sided Greenwood comparison on
$\mathcal R_\delta$. For a positive lower bound, require in addition
$C_G\delta<1$.

\paragraph{Probability of the good event.}
The Dvoretzky--Kiefer--Wolfowitz inequality implies
\[
 \mathbb P(\mathcal A_\delta^c)
 \le
 \mathbb P\left(\sup_{u\le t}|Y(u)/n-y(u)|>\delta p\right)
 \le2\exp(-2n\delta^2p^2).
\]
The inequality also applies to left limits of empirical distribution
functions, as required by $Y(u)=\sum_i\mathbf1\{X_i\ge u\}$.

Set $I(u)=\mathbf1\{Y(u)\ge np/2\}$ and define the predictable
integrand $I(u)/Y(u)$ to be zero when $I(u)=0$.
For the martingale
$Z(v)=\int_0^v I(u)Y(u)^{-1}\,d\bar M_n(u)$,
the jump size is at most $b=2/(np)$ and
$\langle Z\rangle_t\le v=2h/(np)$.
On $\mathcal A_\delta$, $I(u)=1$ throughout $[0,t]$ and
$Z(v)=\widehat\Lambda_n(v)-\Lambda(v)$.
The maximal continuous-time Bernstein--Freedman inequality gives
\begin{align*}
 \mathbb P\left(\sup_{v\le t}|Z(v)|>\delta\right)
 &\le2\exp\left(-\frac{\delta^2}{2(v+b\delta/3)}\right)\\
 &\le2\exp(-C_\Lambda n\delta^2),\qquad
 C_\Lambda=\frac{p}{4(h+1)}.
\end{align*}
Taking a union bound proves
\begin{equation}\label{eq:good_event_prob_S5}
 \mathbb P(\mathcal R_\delta^c)
 \le2\exp(-2n\delta^2p^2)
       +2\exp(-C_\Lambda n\delta^2).
\end{equation}
This argument does not require independence between the risk-set
event and the hazard event.
\end{proof}

\section{Fixed-Cohort Concentration Proofs}\label{app:fixed_concentration}

\paragraph{Sub-Gaussian concentration.}
Write $p=p_{\min}$ and $h=\Lambda(t)$. The risk count $Y(t)$
is binomial with success probability at least $p$, so
$\Prob(Y(t)<np/2)\le e^{-np/8}$. On $Y(t)\ge np/2$ the same
risk lower bound holds at all earlier ages. For the predictable
Nelson--Aalen integrand $\mathbf1\{Y(u)\ge np/2\}/Y(u)$, with
value zero at zero risk, the compensated integral has jump bound
$2/(np)$ and bracket bound $2h/(np)$. On the risk event this
integral equals $\widehat\Lambda_n(t)-h$. For $0<\epsilon\le1$,
Bernstein--Freedman \cite{Freedman1975} implies
\[
 \Prob\bigl(|\widehat\Lambda_n(t)-h|>\epsilon/2,\ Y(t)\ge np/2\bigr)
 \le 2\exp\!\left(-\frac{np\epsilon^2}{16h+8/3}\right).
\]
When $n\epsilon^2\ge4/p$, we have $np/2\ge2$. On the risk event
the no-tie log-product comparison in Appendix~\ref{app:if_proof}
gives $|\widehat S_n(t)-e^{-\widehat\Lambda_n(t)}|\le2/(np)
\le\epsilon/2$. The exponential on nonnegative hazards is
1-Lipschitz, so the union bound yields $3e^{-a n\epsilon^2}$,
where $a=\min\{p/8,p/(16h+8/3)\}$. Set
\[
 c_{\rm KM}=\min\{p\log(2)/4,\ a/2\}.
\]
If $c_{\rm KM}n\epsilon^2\le\log 2$, the stated bound is at least
one. Otherwise $n\epsilon^2>4/p$ and
$3e^{-a n\epsilon^2}\le2e^{-c_{\rm KM}n\epsilon^2}$.
For $\epsilon>1$ the event is empty. This proves
Theorem~\ref{thm:KM-subGaussian}.

\paragraph{Empirical Bernstein concentration.}
Fix $w\ge1$ and $0<\delta<\min\{p/2,1/C_G\}$, with the constants
from Appendix~\ref{app:greenwood_compare}. In addition to
$n\ge n_G(\delta)$, require
\begin{equation}\label{eq:eb_conditions}
 n\ge4/p,\qquad w\le np^2/2,\qquad
 \eta_n(w)\le\delta\sqrt{2w(1-C_G\delta)V(t)/n}.
\end{equation}
These are explicit conditions on deterministic quantities.
For example, under the first two conditions and $w\ge1$,
\[
 \eta_n(w)\le D_\rho w/n,\qquad
 D_\rho=2\sqrt h/p^2+(4h+9)/p.
\]
Indeed, $b\le1/(np)$ in \eqref{eq:full_remainder_tail},
$w/(np)\le1/2$, and $(a+b)^2/2\le a^2+b^2$ applied to
the squared term gives this estimate. A sufficient condition for
the last inequality in \eqref{eq:eb_conditions} is
$n\ge D_\rho^2 w/[2\delta^2(1-C_G\delta)V(t)]$.

Put $\epsilon_n=\sqrt{2w(1-C_G\delta)V(t)/n}$. On $\mathcal R_\delta$
the empirical threshold is at least $\epsilon_n$. Outside the
remainder failure event in \eqref{eq:full_remainder_tail}, the error
exceeding that threshold implies $|L_n(t)|>(1-\delta)\epsilon_n$.
Applying Bernstein to the centered independent influence functions
with variance $V(t)$ and bound $b_\psi$ gives
\[
 \Prob(|L_n(t)|>(1-\delta)\epsilon_n)
 \le 2\exp\!\left(
 -\frac{n(1-\delta)^2\epsilon_n^2}
 {2V(t)+2b_\psi(1-\delta)\epsilon_n/3}\right).
\]
Define
\begin{equation}\label{eq:Rn_def}
 c_\delta(n,w)=
 \frac{(1-\delta)^2(1-C_G\delta)}
      {1+b_\psi(1-\delta)\epsilon_n/(3V(t))}.
\end{equation}
Adding this Bernstein term, the good-event failure probability
\eqref{eq:good_event_prob}, and $6e^{-w}$ gives
\eqref{eq:km_conc_gw}. The radius in that result uses only Greenwood's
variance. The sample condition absorbs the finite-sample remainder,
and its failure probability is included in the tail bound.

\section{Proof of the Regret Bound}\label{app:regret_review}
In this appendix, $t=j-1$ is the decision's information round. Let $\mu_*=\mu_{k^*}$. Write
$\widehat S_k(t^\dagger;t)=\widehat S_{N_k(t)}(t^\dagger,t)$ and
$G_k(t)=\widehat\sigma^2_{N_k(t)}(t^\dagger,t)$.
\subsection{An exponential bound with observed variation}
\label{app:exponential_certificate}
\subsubsection{Construction of the complete-prefix upper bound}
Let $n=n_k^{\mathrm{eff}}(t)$ be the completed prefix length for arm
$k$ at decision time $t$.

For this complete prefix write
\[
 A_k(t)=\sum_{u\le t^\dagger}\frac{\Delta N_{n,k}(u)}{Y_{n,k}(u)},\qquad
 W_k(t)=\sum_{u\le t^\dagger}\frac{\Delta N_{n,k}(u)}{Y_{n,k}(u)^2},
 \quad n=n_k^{\mathrm{eff}}(t),
\]
where the sums are over its observed event times.
Fix $\alpha>1$, $\gamma\ge2$, and use grid weights
\[
 w_0=\tfrac12,\qquad
 w_l=\frac1{4|l|(|l|+1)}\quad(l\in\mathbb Z\setminus\{0\}).
\]
They sum to one. For deterministic $n,t$ define
\[
 x_{t,l}=\gamma\log t+\log(K/w_l),\qquad
 \lambda_{t,n,l}=2^l\sqrt{n\,x_{t,0}},\quad l\in\mathbb Z .
\]
Centering the grid at $\sqrt{n x_{t,0}}$ gives the largest
probability budget to a natural scale for $W_k(t)$ of order $1/n$.

When $W_k(t)>0$, define
\begin{equation}\label{eq:observable_radius}
 a_k(t)=\inf_{l\in\mathbb Z}
 \left\{\frac{\lambda_{t,n,l} W_k(t)}2
                         +\frac{x_{t,l}}{\lambda_{t,n,l}}\right\},
 \qquad
 C_k^{\mathrm{p}}(t)=\exp\!\left(-[A_k(t)-a_k(t)]_+\right).
\end{equation}
If $W_k(t)=0$, including an empty prefix, set $C_k^{\mathrm{p}}(t)=1$.
In particular $W_k(t)=0$ means no event and hence $A_k(t)=0$.

For $W_k(t)>0$ the infimum in \eqref{eq:observable_radius} is
computed by a finite search from $l=0$. Stop in the positive direction
when $\lambda W_k(t)/2$ reaches the smallest candidate found so far.
In the negative direction use $x_{t,l}/\lambda$ as the stopping term.
Both terms are nondecreasing in their respective search directions.

\subsubsection{Probability of underestimation}

Fix an arm and a deterministic complete prefix of size $n$.
Biological age is the time argument in this section.
Write $Y(u)$ for the risk process and $N(u)$ for the observed event
process. Independent censoring gives compensator
$\int_0^uY(v)\,d\Lambda(v)$ \cite{Andersen1993,FlemingHarrington1991}.
Set $f(u)=1/Y(u)$ when $Y(u)>0$ and $f(u)=0$ otherwise.
The left-continuous risk process makes $f$ predictable.
Define
\[
 Z(v)=\int_0^v f(u)\{dN(u)-Y(u)d\Lambda(u)\},\qquad
 W(v)=\int_0^v f(u)^2\,dN(u).
\]

\begin{lemma}\textbf{(One-sided observed-variation bound)}
\label{lem:optional_variation}
For any deterministic $\lambda>0$,
\[
 \mathbb E\exp\{\lambda Z(t^\dagger)-\lambda^2W(t^\dagger)/2\}\le1.
\]
Thus for $x>0$,
\[
 \mathbb P\left(
 \widehat\Lambda_n(t^\dagger)-\Lambda(t^\dagger)>
 \lambda W(t^\dagger)/2+x/\lambda\right)\le e^{-x}.
\]
The second inequality holds even if the risk set is exhausted.
\end{lemma}
\begin{proof}
For $z\ge0$,
$z-z^2/2\le\log(1+z)$, since the derivative of
$\log(1+z)-z+z^2/2$ is $z^2/(1+z)\ge0$.
Consequently $e^{z-z^2/2}-1\le z$.
Put $\theta(u)=\lambda f(u)-\lambda^2f(u)^2/2$.
The counting-process exponential
\[
 \exp\!\left\{\int_0^{t^\dagger}\theta(u)\,dN(u)
 -\int_0^{t^\dagger}(e^{\theta(u)}-1)Y(u)\,d\Lambda(u)\right\}
\]
has expectation at most one as a nonnegative local martingale.
The compensator term is at most
$\lambda\int_0^{t^\dagger} f(u)Y(u)\,d\Lambda(u)$ by the elementary
inequality above. Hence this exponential dominates
$\exp\{\lambda Z(t^\dagger)-\lambda^2W(t^\dagger)/2\}$, proving the first claim.

Since $\widehat\Lambda_n(t^\dagger)=\int_0^{t^\dagger} f\,dN$ and
$\int_0^{t^\dagger} fY\,d\Lambda\le\Lambda(t^\dagger)$, we have
$\widehat\Lambda_n(t^\dagger)-\Lambda(t^\dagger)\le Z(t^\dagger)$
even with an exhausted risk set. Markov's inequality proves the
second claim. 
\end{proof}

\begin{proof}[Proof of Theorem~\ref{thm:certificate}]
For a deterministic prefix define
\[
 Z_n=\widehat\Lambda_n(t^\dagger)-
      \int_0^{t^\dagger}\mathbf1\{Y_n(u)>0\}\,d\Lambda_k(u),\qquad
 W_n=\int_0^{t^\dagger} Y_n(u)^{-2}\,dN_n(u),
\]
taking the integrands as zero at zero risk. The observed-variation
bound above gives
\[
 \mathbb P\left(Z_n>\lambda W_n/2+x/\lambda\right)\le e^{-x},
 \qquad\lambda>0.
\]
Always $\widehat\Lambda_n(t^\dagger)-\Lambda_k(t^\dagger)\le Z_n$,
including when $Y_n(t^\dagger)=0$.
Insert $\lambda=\lambda_{t,n,l}$ and $x=x_{t,l}$, and sum over
$k$, $1\le n\le t$, and $l\in\mathbb Z$, using $\sum_lw_l=1$.
The failure probability is at most $t^{1-\gamma}$.
The simultaneous bound covers all deterministic prefixes before evaluating the randomly selected prefix length.
\end{proof}

\subsection{Deterministic snapshot comparisons}
\label{app:sharp_perturbation}

\begin{lemma}\textbf{(Risk-count bounds)}
\label{lem:risk_count_bounds}
Suppose a base cohort has $n$ complete records through $t^\dagger$ and
target-age risk $s=Y_{\mathrm{base}}(t^\dagger)\ge1$.
Add at most $h$ partial records without changing the base records.
With no event ties,
\begin{align}
 \widehat S_{\mathrm{all}}(t^\dagger)-\widehat S_{\mathrm{base}}(t^\dagger)
 &\le h/s,\label{eq:one_sided_perturbation}\\
 |\widehat S_{\mathrm{all}}(t^\dagger)-\widehat S_{\mathrm{base}}(t^\dagger)|
 &\le 2h/s,\label{eq:two_sided_perturbation}\\
 G_{\mathrm{all}}(t^\dagger)&\le 1/(4s),\qquad
 W_{\mathrm{base}}(t^\dagger)\le 1/s.\label{eq:risk_greenwood}
\end{align}
These comparisons hold pathwise, even if the extra records depend
on the history.
\end{lemma}
\begin{proof}
There is almost surely no event exactly at the fixed target age.
The pre-event base risk counts are distinct integers at least $s+1$:
each event reduces the risk count, and $s$ base patients remain
at risk through the target. Thus
\[
 \sum_{\text{base events}}\frac1{Y(Y-1)}
 \le\sum_{y=s+1}^n\frac1{y(y-1)}\le\frac1s,\qquad
 W_{\mathrm{base}}\le\sum_{y=s+1}^n y^{-2}\le\frac1s .
\]
At a base event, adding records increases its risk count by at most
$h$. The derivative of $\log(1-1/y)$ is $1/[y(y-1)]$,
so the total increase in the base-event log factors is at most
$h/s$. The extra events only decrease the log product. This proves
the one-sided bound for the log survival difference, and hence
\eqref{eq:one_sided_perturbation} because the exponential is
1-Lipschitz on $(-\infty,0]$.
There are at most $h$ extra events, each with at least $s+1$ at risk,
so their log losses are at most $h/s$ in total.
This also proves \eqref{eq:two_sided_perturbation}.

For the full snapshot, let $q_i=1/Y_i$ at its event times and
write $S^\flat=\prod_i(1-q_i)>0$. Every $q_i\le1/s$.
By the nonnegative product expansion,
\[
 \sum_i\frac{q_i}{1-q_i}
 \le\prod_i\left(1+\frac{q_i}{1-q_i}\right)-1
 =1/S^\flat-1.
\]
Hence
\[
 G_{\mathrm{all}}
 =(S^\flat)^2\sum_i\frac{q_i^2}{1-q_i}
 \le\frac{S^\flat(1-S^\flat)}s\le\frac1{4s}.
\]
This controls the full-snapshot Greenwood term directly.
\end{proof}

\subsection{Decision-time events and the index width}
\label{app:sharp_regret}

Write $V_k=V_k(t^\dagger)$ in this section. Let
$\kappa=-\log(1-r)$ and put
\begin{align}
 d_t&=\left\lceil\gamma\log t/\kappa\right\rceil,
 &h_t&=\max\{\lceil t^\dagger\rceil,d_t\},\\
 Z_T&=\gamma\log T+\log(4K),
 &L_h&=1/\kappa+(\lceil t^\dagger\rceil+1)/Z_2,\\
 M&=4/p+14/(3p^2).
\end{align}
Then $h_t\le L_h Z_T$ for $2\le t\le T$.
For $\Lambda_k(t^\dagger)>0$ define the proof constants
\begin{align}
 l_k&=\operatorname{round}\left(\log_2\sqrt{2/(M\Lambda_k(t^\dagger))}\right),
 &d_k&=\log(w_0/w_{l_k}),\\
 C_k^{\rm width}
 &=\frac1p\left[\frac2{\sqrt p}+
  \max\left\{4\sqrt{\alpha/p},\,
   \sqrt{M/2}\left(2\sqrt2+\frac{\sqrt2d_k}{x_{2,0}}\right)\right\}\right],
 \label{eq:sharpened_regret_constants}\\
 D&=\max\left\{1,\frac{4/3+2L_h+\sqrt{2\alpha(28/3+4L_h)}}p\right\}.
\end{align}
\paragraph{Reporting.}
The record through age $t^\dagger$ is available by round
$i+\max\{D_i,\lceil t^\dagger\rceil\}$ since by then the patient is in
the observation pool and has reached age $t^\dagger$. It may be
available sooner if a terminal outcome has been reported.
For $\mathcal D_t=\{\max_{i\le t}D_i\le d_t\}$, the geometric tail
gives $\mathbb P(\mathcal D_t^c)\le t(1-r)^{d_t}\le t^{1-\gamma}$.
On $\mathcal D_t$, every patient $i\le t-h_t$ has a complete
target record by $t$, since $i+\max\{D_i,\lceil t^\dagger\rceil\}\le t$.
For every arm, those old patients are part of its complete prefix.
Consequently
\[
 n_k^{\mathrm{eff}}(t)\ge N_k(t)-h_t.
\]
The full realtime snapshot contains that prefix, and at most $h_t$
more records. This fact does not require the ready patients to
arrive in order. The prefix ends at the first missing record.

\paragraph{Risk count and a two-sided hazard bound for the proof.}
Let $x_t=\gamma\log t+\log(4K)$.
For a fixed complete cohort of size $n$, its target risk count
$Y_n(t^\dagger)$ is binomial with success probability at least $p$.
Thus
\[
 \mathbb P(Y_n(t^\dagger)<np/2)\le e^{-np/8}.
\]
This Chernoff bound is applied only at the fixed target age.
Monotonicity of the risk count gives the lower bound at earlier
ages on the same event.

For an age-scale martingale calculation, use the predictable
integrand
\[
 \frac{\mathbf 1\{Y_n(u)\ge np/2\}}{Y_n(u)},
\]
interpreted as zero outside the indicated set. Its compensated integral has jumps at
most $2/(np)$ and bracket at most $2\Lambda_k(t^\dagger)/(np)$.
The two-sided Bernstein--Freedman bound \cite{Freedman1975} gives,
outside probability $2e^{-x_t}$, magnitude at most
\[
 w_{k,t}(n)=2\sqrt{\frac{\Lambda_k(t^\dagger)x_t}{np}}+\frac{4x_t}{3np}.
\]
On $Y_n(t^\dagger)\ge np/2$, the integral is exactly
$\widehat\Lambda_n(t^\dagger)-\Lambda(t^\dagger)$.
For $n\ge8x_t/p$, the risk-count failure is at most $e^{-x_t}$.
A union bound over $K$ arms and all deterministic $n\le t$ shows
that both bounds hold for all such prefixes with failure probability
at most $(3/4)t^{1-\gamma}$.

Let $\mathcal H_t$ intersect this event with $\mathcal D_t$ and the
certificate event of Theorem~\ref{thm:certificate}.
Then $\mathbb P(\mathcal H_t^c)\le3t^{1-\gamma}$.
Note that $\mathcal H_t$ is purely an analytical event constructed for the proof. It uses unconditional probabilities and pathwise bounds based on latent records.

\paragraph{A variance-dependent upper bound on the index.}
On $\mathcal H_t$, let $n=n_k^{\mathrm{eff}}(t)\ge8Z_T/p$ and
$s=Y_n(t^\dagger)\ge np/2$. Write $\Lambda=\Lambda_k(t^\dagger)$,
$V=V_k$ and $z=Z_T/n$. The predictable variance bound for this
arm gives
\[
 |A_k(t)-\Lambda|\le 2\sqrt{\Lambda z/p}+4z/(3p).
\]
Since $W_k(t)\le A_k(t)/s$ and $2\sqrt{\Lambda z/p}\le\Lambda+z/p$,
\begin{equation}\label{eq:variance_observed}
 W_k(t)\le \frac{4\Lambda}{np}+\frac{14z}{3np^2}.
\end{equation}
At a prefix event the full risk set is at least as large as the
prefix risk set, and the full survival estimate is at most one.
Since each risk at an event is at least $s+1\ge2$, the base
event contributions to Greenwood sum to at most $2W_k(t)$.
There are at most $h_t$ extra failures, each contributing at
most $1/[s(s+1)]$. It follows that
\[
 G_k(t)\le 2W_k(t)+\frac{h_t}{s(s+1)}.
\]
Also $\widehat S_{\mathrm{prefix}}\le e^{-A_k(t)}$ and
\eqref{eq:one_sided_perturbation} gives
$\widehat S_k(t^\dagger;t)-\mu_k\le |A_k(t)-\Lambda|+h_t/s$.
With $h_t\le L_h Z_T$ and $\log t\le Z_T$, these estimates give
\[
 U_k^0(t)-\mu_k\le
 \left(2/\sqrt p+4\sqrt{\alpha/p}\right)\sqrt{\Lambda z}
 +\left[\frac{4/3+2L_h}{p}
        +\frac{\sqrt{2\alpha(28/3+4L_h)}}p\right]z.
\]

For the prefix upper bound, if $\Lambda\le z$ then
$C_k^{\mathrm p}(t)-\mu_k\le 1-e^{-\Lambda}\le z$.
Otherwise, \eqref{eq:variance_observed} gives $W_k(t)\le M\Lambda/n$.
For $W_k(t)>0$ use the grid candidate $l_k$ from
\eqref{eq:sharpened_regret_constants}, with
$v_k=2^{l_k}\sqrt{M\Lambda/2}\in[2^{-1/2},2^{1/2}]$ and
$\lambda_{t,n,l_k}=v_k\sqrt{2n x_{t,0}/(M\Lambda)}$. Then
\begin{align*}
 a_k(t)&\le \sqrt{\frac{M\Lambda}{2n}}
 \left[(v_k+v_k^{-1})\sqrt{x_{t,0}}+
                    \frac{d_k}{v_k\sqrt{x_{t,0}}}\right]\\
 &\le \sqrt{\frac{M\Lambda z}{2}}
           \left(2\sqrt2+\frac{\sqrt2 d_k}{x_{2,0}}\right).
\end{align*}
The same inequality holds at $W_k(t)=0$ if one defines $a_k(t)=0$
there, since $A_k(t)=0$ and the prefix certificate is one.
The positive-part map and $u\mapsto e^{-u}$ on $[0,\infty)$
are 1-Lipschitz. Consequently
$C_k^{\mathrm p}(t)-\mu_k\le a_k(t)+|A_k(t)-\Lambda|$.
Finally $V=\mu_k^2\int_0^{t^\dagger} y_k(u)^{-1}\,d\Lambda_k(u)
\ge p^2\Lambda$. Thus
the two bounds imply
\begin{equation}\label{eq:pointwise_index_width}
 \max\{U_k^0(t),C_k^{\mathrm p}(t)\}-\mu_k\le C_k^{\rm width}\sqrt{V_k Z_T/n}+D Z_T/n.
\end{equation}
For an arm with $\Lambda=0$, no failures occur through $t^\dagger$ and
$U_k^0(t)=C_k^{\mathrm p}(t)=1=\mu_k$ on the stated event.

\paragraph{A lower bound on the original index.}
For $n\ge 8Z_T/p$, the target risk $s$ in the complete prefix is
positive. At the prefix failures the distinct pre-jump risk counts
are integers at least $s+1$. The log-product correction satisfies
$0\le-\log\widehat S_{\rm prefix}-A_k(t)\le1/s$ because the sum of
$1/[y(y-1)]$ over these counts is at most $1/s$.
Adding the other reports only increases the product factors at
prefix failures. At most $h_t$ new failures occur and their log
factors are no smaller than $-1/s$ each. Since the exponential
on nonpositive log survival values is 1-Lipschitz and the original
bonus is nonnegative, it follows that
\begin{align}
 \mu_k-U_k^0(t)
 &\le |A_k(t)-\Lambda_k(t^\dagger)|+(1+h_t)/s\notag\\
 &\le a_-\sqrt{V_k Z_T/n}+D_- Z_T/n,
 \label{eq:wait_index_lower}
\end{align}
where $a_-=2/p^{3/2}$ and $D_-=(4/3+2+2L_h)/p$.
The inequalities $n\ge8Z_T/p$, $s\ge np/2$, $Z_T\ge1$ and
$h_t\le L_h Z_T$
ensure that all these estimates are valid.

\subsubsection{Concentration for the realtime KM estimate}
\label{app:bandit_concentration_proof}
\paragraph{Proof of Theorem~\ref{thm:bandit_concentration}.}
Write $\kappa=-\ln(1-r)$ and set
\[
 d_T^{(\alpha)}=\left\lceil x_T/\kappa\right\rceil,\qquad
 h_T^{(\alpha)}=\max\{\lceil t^\dagger\rceil,d_T^{(\alpha)}\}.
\]
The probability that any of the first $T$ reporting delays exceeds
$d_T^{(\alpha)}$ is at most $T e^{-x_T}$. On the complementary
event, the target record of every patient enrolled through
$T-h_T^{(\alpha)}$ is complete by $T$. This gives
$n_k^{\mathrm{eff}}(T)\ge N_k(T)-h_T^{(\alpha)}$ and leaves at most
$h_T^{(\alpha)}$ additional records in the realtime snapshot of an arm.

For a deterministic complete prefix of size $n\ge8x_T/p$, the
probability that the sample risk at $t^\dagger$ is below $np/2$ is
at most $e^{-np/8}\le e^{-x_T}$. By
Theorem~\ref{thm:KM-subGaussian}, the fixed-prefix KM error exceeds
$\sqrt{V_{\mathrm{sub},k}x_T/n}$ with probability at most
$2e^{-x_T}$. If $\Lambda_k(t^\dagger)=0$, the prefix KM estimate
equals one and the error is zero; the same constant formula is
well defined. A union bound over $K$ arms and $1\le n\le T$ costs
at most $3KT e^{-x_T}$. On its complementary event the bounds
hold for the actual prefix lengths meeting the sample condition.

Let $s\ge np/2$ be the number at risk at the target in such a
prefix. In the log-KM product, adding the extra records increases
the terms at original failure times by a total of at most
$h_T^{(\alpha)}/s$. It can introduce new negative terms of total
magnitude at most $h_T^{(\alpha)}/s$ by the same risk-count
comparison. Hence the absolute change in log survival is at most
$h_T^{(\alpha)}/s$. Exponentiation on the nonpositive half-line is
1-Lipschitz, so the KM change is at most $2h_T^{(\alpha)}/(np)$.
Combining the fixed-prefix and reporting bounds yields
\eqref{eq:bandit_conc} with failure probability at most
$(3K+1)T e^{-x_T}=T^{-\alpha}$.

\subsection{Proof of the regret theorem}\label{app:three_events}
Put $A_k=C_k^{\rm width}+a_-$, $D_{\rm r}=D+\max\{c,D_-\}$,
$L_0=8/p$ and $\Delta_{\min}=\min_{k\ne k^*}\Delta_k$.
For $\Lambda_k(t^\dagger)=0$ use $C_k^{\rm width}=0$.
Write $Z=Z_T$, $h=h_T$, $\ell=\lceil L_0Z\rceil$, and define
\[
 \mathcal B_k(m)=
 \begin{cases}
 A_k\sqrt{V_k Z/(m-h)}+D_{\rm r} Z/(m-h), &m-h\ge \ell,\\
 +\infty, &m-h<\ell .
 \end{cases}
\]
The common good event $\mathcal H_t$ from the preceding subsection
has failure probability at most $3t^{1-\gamma}$.
On this event the effective sample size for arm $k$ is at least
$N_k(t)-h$. By \eqref{eq:pointwise_index_width},
\eqref{eq:wait_index_lower}, and $\ln t/N_k(t)\le Z/n$
when $n=n_k^{\rm eff}(t)>0$, each of the following is at most
$\mathcal B_k(N_k(t))$:
\begin{equation}\label{eq:review_three_widths}
 C_k^{\mathrm p}(t)-\mu_k,\qquad
 U_k^0(t)+c\ln t/N_k(t)-\mu_k,\qquad
 C_k^{\mathrm p}(t)-U_k^0(t).
\end{equation}
The claim holds trivially for an infinite width.

\paragraph{The three events.}
For decision $j=t+1$, let $\mathcal E_1(j)$ denote
$C_{k^*}^{\mathrm p}(t)<\mu_*$ and let $\mathcal E_2(j)$ denote
a violation of \eqref{eq:review_three_widths} for any arm.
The regret contribution of their union is at most $3\ln T$, since
both events are excluded on $\mathcal H_t$, a decision has regret
at most one, and
$\sum_{t=J_0}^{T-1}t^{1-\gamma}\le\sum_{t=2}^{T-1}t^{-1}\le\ln T$.
This proves Lemma~\ref{lem:est_fail} and counts every affected decision,
even when several decisions use the same optimal-arm data.

Suppose both events fail to occur. Then the candidate $k=k_j$
has $C_k^{\mathrm p}(t)\ge\mu_*$. If the first condition in
\eqref{eq:review_rule} fails, a suboptimal selected arm $b=b_j$
satisfies $\Delta_b\le \mathcal B_b(N_b(t))$ by the second
width in \eqref{eq:review_three_widths}. If review executes,
the selected arm is $k$ and its gap is at most
$\mathcal B_k(N_k(t))$ by the first width. If the first condition
holds but the waiting condition fails, then $a_j=b$ and
$U_b^0(t)\ge U_k^0(t)$. In this case,
\begin{equation}
 \Delta_b\le
   \bigl[C_k^{\mathrm p}(t)-U_k^0(t)\bigr]
      +\bigl[U_b^0(t)-\mu_b\bigr].
 \label{eq:blocked_review_decomp}
\end{equation}
If the second term is at least $\Delta_b/2$, charge the decision
to the selected arm. Otherwise the first is at least $\Delta_b/2$,
and we charge the decision to the candidate.
Thus every suboptimal decision outside $\mathcal E_1\cup\mathcal E_2$
has a width at least half the selected arm's gap, the third event
$\mathcal E_3(j)$. For an unblocked decision, the event uses the width
of the selected arm. For a blocked review, it permits the width of
either the selected arm or the candidate. The two charge rules above
assign each such decision to just one arm. This proves
Lemma~\ref{lem:decomp} after counting the initialization pulls.

\paragraph{Counting by pulls of the selected arm.}
For $0<g\le1$ put
\[
 q_k(g)=\left\lceil Z\max\left\{L_0,\frac{4A_k^2 V_k}{g^2},
                                     \frac{2D_{\rm r}}g\right\}\right\rceil .
\]
If $m>h+q_k(g)$, then $\mathcal B_k(m)<g$.
A selected-arm charge to a suboptimal arm $b$ therefore requires
$N_b(t)\le h+q_b(\Delta_b/2)$. At decisions where $a_j=b$,
each pre-decision count occurs at most once. Counting
initialization separately, the selected-arm exploration contribution
is bounded by
\begin{equation}\label{eq:selected_review_cost}
 B_{\mathrm{sel}}(T)=\sum_{b\ne k^*}\Delta_b
       \bigl[h+q_b(\Delta_b/2)+1\bigr].
\end{equation}
Initialization adds $n_{\mathrm{init}}\sum_{b\ne k^*}\Delta_b$.

\paragraph{Counting blocked reviews by the candidate's latest pull.}
A candidate charge requires $t-\ell_k(t)<L$, where $\ell_k(t)$
is the last pull of that candidate. On such a decision,
$\Delta_b\le\min\{1,2\mathcal B_k(N_k(t))\}$ by
\eqref{eq:blocked_review_decomp} and the charge rule.
Also $\mathcal B_k(N_k(t))\ge\Delta_k$ since
$C_k^{\mathrm p}(t)\ge\mu_*$. Therefore, writing
$g_k=\max\{\Delta_k,\Delta_{\min}/2\}$, a charge to candidate $k$
requires $N_k(t)\le h+q_k(g_k)$, even if $k=k^*$.

For each arm $k$ and count $m\ge1$, fix the round in which its
$m$th pull occurred. While $N_k(t)=m$, that same pull is the latest.
Only the first $L$ candidate-decision times after that pull can
satisfy the blocking inequality. Thus each $(k,m)$ accounts
for at most $L$ charges, including when the identity of the
candidate changes across decisions. The total contribution is at most
\begin{equation}\label{eq:blocked_review_cost}
 B_{\mathrm{wait}}(T)=L\sum_{k=1}^K\sum_{m=1}^{h+q_k(g_k)}
        \min\{1,\,2\mathcal B_k(m)\}.
\end{equation}
Equations~\eqref{eq:selected_review_cost} and
\eqref{eq:blocked_review_cost} give the exploration bound in
Lemma~\ref{lem:explore}. Adding initial pulls and the two
decision-time errors gives the finite-horizon regret bound
before simplifying constants.

\paragraph{Simplifying the regret bound.}
For completeness put $z_0=\gamma+\ln(4K)/\ln 2$ so that
$Z\le z_0\ln T$, and define
\begin{align*}
 u_k(g)&=L_0+1+4A_k^2V_k/g^2+2D_{\rm r}/g,\\
 E_k(g)&=L_h+L_0+1+4A_k\sqrt{V_k(L_0+1)}+4D_{\rm r}\\
 &\quad+2D_{\rm r}\bigl[1+\ln(u_k(g)/L_0)\bigr].
\end{align*}
To bound the inner sum of \eqref{eq:blocked_review_cost}, there
are at most $h+\ell$ counts before the finite width applies.
For the remaining counts, sum $(m-h)^{-1/2}$ and $(m-h)^{-1}$
from $\ell$ to $q_k(g)$. Their sums are at most
$2\sqrt{q_k(g)}$ and $\ell^{-1}+\ln(q_k(g)/\ell)$, respectively.
Since $q_k(g)\le Z u_k(g)$ and $\ell\ge L_0 Z$, these estimates and
$2\sqrt{xy}\le x+y$ give
\[
 \sum_{m=1}^{h+q_k(g)}\min\{1,\,2\mathcal B_k(m)\}
 \le Z\left[10A_k^2 V_k/g+E_k(g)\right].
\]
In particular the harmonic sum is bounded in $T$ for a fixed instance.
The selected-arm sum together with initialization is at most
$Z\sum_{b\ne k^*}[16A_b^2 V_b/\Delta_b+4D_{\rm r}
       +(n_{\mathrm{init}}+L_h+L_0+2)\Delta_b]$.
For a suboptimal candidate $g_k\ge\Delta_k$, so its variance term
can be combined with the corresponding selected-arm term.
For the optimal candidate $g_{k^*}=\Delta_{\min}/2$, giving the
additional $L V_{k^*}/\Delta_{\min}$ term. In particular
\eqref{eq:sharpened_regret} holds with
\begin{align*}
 C &=z_0(16+10L)\max_k A_k^2,\\
 C_{\mathrm{rem}}
 &=3+z_0\Bigg[
   \sum_{b\ne k^*}\bigl(4D_{\rm r}
        +(n_{\mathrm{init}}+L_h+L_0+2)\Delta_b\bigr)\\
 &\qquad\qquad
   +L\sum_k E_k(g_k)+
          \frac{20L A_{k^*}^2V_{k^*}}{\Delta_{\min}}
   \Bigg].
\end{align*}
The constants depend on the model and fixed review parameters.
The optimal-candidate term and the logarithmic gap factors in
$E_k(g_k)$ quantify the additional cost of the waiting condition.

\section{Persistent Exploration Guarantee}\label{app:persistent_exploration}
\begin{lemma}\textbf{(No finite upper bound on arm pulls)}
\label{lem:lowerbound}
If $\max_k\mu_k<1$, every arm of the review policy is sampled
infinitely often almost surely.
\end{lemma}
\begin{proof}
Fix $\max_k\mu_k<z<1$. On $\mathcal H_t$, if
$\max_k C_k^{\mathrm p}(t)\ge z$, one of three things occurs.
If the first review condition fails, the selected original index plus
its review allowance is at least $z$. If the review executes, the
selected certificate is at least $z$. In either case the selected
arm has width at least $z-\mu_k$. If the review is blocked, the
candidate has certificate at least $z$, and that same width condition
holds for the candidate. Count the first two cases by selected-arm
pulls and the last by at most $L$ decision times per candidate pull,
as in Section~\ref{app:three_events}. Because $z-\mu_k$ is positive
for every arm, the expected number of these rounds is $O(\ln T)$,
including confidence failures. Markov's inequality and
Borel--Cantelli along dyadic horizons show that their fraction
converges to zero almost surely on that subsequence.

If some arm is sampled only finitely often, eventually every record
in its fixed prefix is complete. The radius in
\eqref{eq:observable_radius} is at least $\sqrt{2W_k(t)x_{t,0}}$,
by the arithmetic--geometric mean inequality and $x_{t,l}\ge x_{t,0}$.
Its certificate therefore eventually equals one, also when $W_k=0$.
Then $\max_k C_k^{\mathrm p}(t)\ge z$ at every sufficiently late
decision, a contradiction.
\end{proof}

\section{Asymptotic Normality under Random Sample Size}
\label{app:asym_normal}
\begin{theorem}
Fix an arm $k$. In the reporting model of Section~\ref{sec:setup},
suppose a deterministic sequence $a_T\to\infty$ satisfies
$N_k(T)/a_T\to1$ in probability and $\log T/\sqrt{a_T}\to0$.
Then the realtime estimator satisfies
\[
 \sqrt{N_k(T)}\bigl(\widehat S_{N_k(T)}(t^\dagger,T)-S_k(t^\dagger)\bigr)
 \Rightarrow \mathcal N(0,V_k(t^\dagger)).
\]
\end{theorem}
\begin{proof}
Write $n=n_k^{\mathrm{eff}}(T)$. The reporting bound gives
$0\le N_k(T)-n\le h_T=O(\log T)$ with probability tending to one.
Hence $n/a_T\to1$ in probability. For the centered bounded influence
functions, Kolmogorov's maximal inequality applied on either side of
$\lfloor a_T\rfloor$ shows that the difference of the sums at $n$
and $\lfloor a_T\rfloor$, divided by $\sqrt{a_T}$, tends to zero
in probability. Indeed, on $|n-\lfloor a_T\rfloor|\le\epsilon a_T$ the
maximal inequality bounds the failure probability at any fixed
threshold by a constant times $\epsilon$. Lemma~\ref{lem:IF_rep}
then gives the prefix limit by the ordinary i.i.d.\ central limit
theorem. The almost sure eventual remainder bound applies along the
prefixes since $n\to\infty$ in probability.

The strong law for the target risk count of the latent prefix gives
$Y_n(t^\dagger)\ge np/2$ with probability tending to one.
By \eqref{eq:two_sided_perturbation}, the snapshot difference is at most
$2h_T/Y_n(t^\dagger)$. Multiplication by $\sqrt{N_k(T)}$ yields a
quantity tending to zero in probability. Slutsky's theorem proves
the stated result.
\end{proof}

\section{Properties of the Greenwood Variance in the Uncensored Limit}\label{app:gw_nocensor}
Without censoring ($C_i \equiv \infty$) and with $t^\dagger$ equal to a fixed, fully observed target age, the KM estimator reduces to the empirical proportion $\widehat S_{N_k(T)}(t^\dagger, T) = n_k(t^\dagger,T)/N_k(T)$ and the Greenwood formula simplifies to
\begin{equation}\label{eq:gw_nocensor}
  \widehat{\sigma}_{N_k(T)}^2(t^\dagger, T) = \frac{\widehat S_{N_k(T)}(t^\dagger, T)\bigl(1 - \widehat S_{N_k(T)}(t^\dagger, T)\bigr)}{N_k(T)},
\end{equation}
which is the plug-in binomial variance $\widehat S(1-\widehat S)/n$. This confirms that our variance-adaptive confidence interval reduces to the classical Bernstein-type bound in the absence of censoring, and the improvement from using the Greenwood formula is purely due to the censoring structure.


\clearpage
\begin{thebibliography}{16}
\bibitem{NIPS2016_0bf727e9}
J. D. Abernethy, K. Amin, and R. Zhu,
Threshold bandits, with and without censored feedback,
in Advances in Neural Information Processing Systems,
D. Lee, M. Sugiyama, U. Luxburg, I. Guyon, and R. Garnett, eds.,
vol. 29, Curran Associates, Inc., 2016.
\url{https://proceedings.neurips.cc/paper_files/paper/2016/file/0bf727e907c5fc9d5356f11e4c45d613-Paper.pdf}.
\bibitem{Andersen1993}
P. K. Andersen, {\O}. Borgan, R. D. Gill, and N. Keiding,
Statistical Models Based on Counting Processes, Springer, 1993.
\bibitem{auer2002finite}
P. Auer, N. Cesa-Bianchi, and P. Fischer,
Finite-time analysis of the multiarmed bandit problem,
Machine Learning, 47 (2002), pp. 235--256.
\bibitem{bubeck2012regret}
S. Bubeck and N. Cesa-Bianchi,
Regret analysis of stochastic and nonstochastic multi-armed bandit problems,
Foundations and Trends in Machine Learning, 5 (2012), pp. 1--122.
\bibitem{Chapelle2014}
O. Chapelle, Modeling delayed feedback in display advertising,
in Proceedings of the 20th ACM SIGKDD International Conference
on Knowledge Discovery and Data Mining, 2014, pp. 1097--1105.
\bibitem{FlemingHarrington1991}
T. R. Fleming and D. P. Harrington,
Counting Processes and Survival Analysis, John Wiley \& Sons, 1991.
\bibitem{Gill1983}
R. D. Gill,
Large sample behaviour of the product-limit estimator on the whole line,
Annals of Statistics, 11 (1983), pp. 49--58.
\bibitem{goel2010understanding}
M. K. Goel, P. Khanna, and J. Kishore,
Understanding survival analysis: Kaplan--Meier estimate,
International Journal of Ayurveda Research, 1 (2010), p. 274.
\bibitem{Greenwood1926}
M. Greenwood, The natural duration of cancer,
Reports on Public Health and Medical Subjects, 33 (1926), pp. 1--26.
\bibitem{Kaplan1958}
E. L. Kaplan and P. Meier,
Nonparametric estimation from incomplete observations,
Journal of the American Statistical Association, 53 (1958), pp. 457--481.
\bibitem{lai1985asymptotically}
T. L. Lai and H. Robbins, Asymptotically efficient adaptive allocation rules,
Advances in Applied Mathematics, 6 (1985), pp. 4--22.
\bibitem{villar2015multi}
S. S. Villar, J. Bowden, and J. Wason,
Multi-armed bandit models for the optimal design of clinical trials:
Benefits and challenges, Statistical Science, 30 (2015), pp. 199--215.
\bibitem{wu2022partial}
H. Wu and S. Wager, Partial likelihood Thompson sampling,
in Proceedings of the 38th Conference on Uncertainty in Artificial Intelligence,
vol. 180 of Proceedings of Machine Learning Research,
PMLR, 2022, pp. 2138--2147.
\bibitem{Audibert2009}
J.-Y. Audibert, R. Munos, and C. Szepesv{\'a}ri,
Exploration--exploitation tradeoff using variance estimates in
multi-armed bandits, Theoretical Computer Science, 410 (2009),
pp. 1876--1902.
\url{https://imagine.enpc.fr/publications/papers/TCS08.pdf}.
\bibitem{Freedman1975}
D. A. Freedman, On tail probabilities for martingales,
The Annals of Probability, 3 (1975), pp. 100--118.
\bibitem{Massart1990}
P. Massart, The tight constant in the Dvoretzky--Kiefer--Wolfowitz
inequality, The Annals of Probability, 18 (1990), pp. 1269--1283.
\end{thebibliography}
\end{document}